\documentclass[twoside,11pt]{article}
\pdfoutput=1
\usepackage{geometry}
\usepackage[colorlinks=true]{hyperref}

\usepackage{amsmath}
\usepackage{amsfonts}
\usepackage{amssymb}
\usepackage{graphicx, subfigure, fink, grffile, placeins}
\usepackage{color}
\usepackage{url}
\usepackage{mdwlist}
\usepackage{euscript}

\usepackage{alex}
\usepackage{algorithm,algorithmic}
\usepackage{booktabs}
\usepackage{multirow}
\usepackage{xspace}

\usepackage{tikz}
\usetikzlibrary{calc,positioning,arrows}
\usetikzlibrary{automata,positioning}
\usetikzlibrary{fit}         
\usetikzlibrary{backgrounds} 
\usetikzlibrary{positioning}
\usetikzlibrary{shadows}
\usepgflibrary{shapes}

\tikzstyle{observed}=[circle, thick, minimum size=0.9cm, draw=black!100, fill=black!20]
\tikzstyle{latent}=[circle, thick, minimum size=0.9cm, draw=black!80]
\tikzstyle{plate}=[rectangle, thick, inner sep=0.25cm, draw=black!100]
\tikzstyle{shadeplate}=[rectangle, thick, inner sep=0.4cm, draw=black!100]
\tikzstyle{table}=[circle,fill=blue!20,draw=black!100,inner sep=1pt, minimum size=30pt]
\tikzstyle{client}=[rectangle,fill=blue!20,draw=black!100,inner sep=1pt, minimum size=12pt]

\newcommand{\ourmodel}{\rm PACO\xspace}

\newcommand{\textdata}{n}

\newcommand{\data}{\mathbf{R}}

\newcommand{\sindex}{\ell}
\newcommand{\clindex}{a}
\newcommand{\clindexm}{b}
\newcommand{\sindexp}{ {(\sindex)} }
\newcommand{\Sfn}{\mathcal{S}}
\newcommand{\template}{\mathbf{T}}
\newcommand{\uclusters}{\mathbf{c}}
\newcommand{\mclusters}{\mathbf{d}}

\newcommand{\nstencils}{S}

\newcommand{\Sfnp}{\Sfn\left(\template, \uclusters, \mclusters\right)}
\newcommand{\Sfnpi}{\Sfn\left(\template^\sindexp, \uclusters^\sindexp, \mclusters^\sindexp\right)}

\newcommand{\email}[1]{\href{mailto:#1}{\tt #1}}

\definecolor{darkblue}{rgb}{0.0,0.0,0.4}

\hypersetup{
  colorlinks = true,
  urlcolor = darkblue
}

\title{Explaining reviews and ratings with PACO: \\
  Poisson Additive Co-Clustering}

\author{
Chao-Yuan Wu\thanks{These authors contributed equally.} \\
Department of Computer Science\\
University of Texas at Austin, Austin, TX\\
\email{cywu@cs.utexas.edu}\\
\and
Alex Beutel$^*$ \\
Computer Science Department\\
Carnegie Mellon University, Pittsburgh, PA\\
\email{abeutel@cs.cmu.edu}\\
\and
Amr Ahmed \\
Google Strategic Technologies, Mountain View, CA\\
\email{amra@google.com}
\and
Alexander J.\ Smola \\
Carnegie Mellon University, Pittsburgh, PA\\
Marianas Labs, Mountain View, CA\\
\email{alex@smola.org}
}

\begin{document}

\maketitle

\begin{abstract}
  Understanding a user's \emph{motivations} provides valuable
  information beyond the ability to recommend items. Quite often this
  can be accomplished by perusing both ratings and review texts, since
  it is the latter where the reasoning for specific preferences is
  explicitly expressed. 

  Unfortunately matrix factorization approaches to recommendation
  result in large, complex models that are difficult to interpret and
  give recommendations that are hard to clearly explain to users.  In
  contrast, in this paper, we attack this problem through succinct
  additive co-clustering.  
  We devise a novel Bayesian technique for summing co-clusterings of Poisson
  distributions.
  With this novel technique we propose a new Bayesian model for joint
  collaborative filtering of ratings and text reviews through a sum of simple
  co-clusterings.
  The simple structure of our model yields easily
  interpretable recommendations.  Even with a simple, succinct
  structure, our model outperforms competitors in terms of predicting ratings
  with reviews.
\end{abstract}

\section{Introduction} 
\label{sec:Introduction}

Recommender systems often serve a dual purpose --- they are expected
to generate suggestions that users might like, while simultaneously
being able to \emph{explain} why a certain recommendation was
made. This increases a user's confidence in a recommender system and
it offers valuable insight for debugging a malfunctioning model. 

Matrix factorization \cite{Koren2009MatrixFactorization} accomplishes
this goal only to a limited extent, since it maps all users and movies
into a rather low-dimensional space, where objects are compared by the
extent of overlap they have in terms of their inner product. This
limits attempts to understand the model to principal component analysis
and nearest neighbor queries for specific instances. 

On the other hand, in many cases users are actually happy to provide
explicit justification for their preferences in the form of written
reviews, albeit in \emph{free text form} rather than as categorized
feedback. They offer immediate insights into the reasoning, provided
that we are able to capture this reasoning in the form of a model for
the text inherent in the reviews. JMARS \cite{diao2014jointly}
exploited this insight by designing a topic model to capture reviews
and ratings jointly, thus offering one of the first works to infer
both topics and sentiments without requiring explicit aspect ratings.

A challenge in these approaches is that the model 
must fit a language model to a rather messy, high dimensional embedding
of users and items.
In the context of recommender systems
ACCAMS \cite{accams} addressed this problem by introducing a novel
additive co-clustering model for matrix completion. This approach yielded
excellent predictive accuracy while providing a well-structured, parsimonious
model.
Because the model heavily relied on backfitting and an additive
representation of a regression model, 
it is not possible to combine it with multinomial language models,
i.e.\ a simple bag-of-words representation, since
probabilities are not additive: They need to be normalized to 1. 

We address this problem by introducing a novel \emph{additive} language
description in the form of a sum of \emph{Poisson} distributions rather than a
Binomial distribution. 
This strategy allows us to use backfitting for documents rather than just in a
regression setting, and enables a wide variety of new applications. 
This is possible because the Poisson distribution is
closed under addition. This means that sums of Poisson random variables remain
Poisson. This property also applies to mixtures of Poisson random variables,
i.e.\ the occurrence of multiple words.

\begin{figure}[tb!]
\centering
\subfigure[Amazon fine foods]{\label{fig:joint_a}\includegraphics[width=0.32\textwidth]{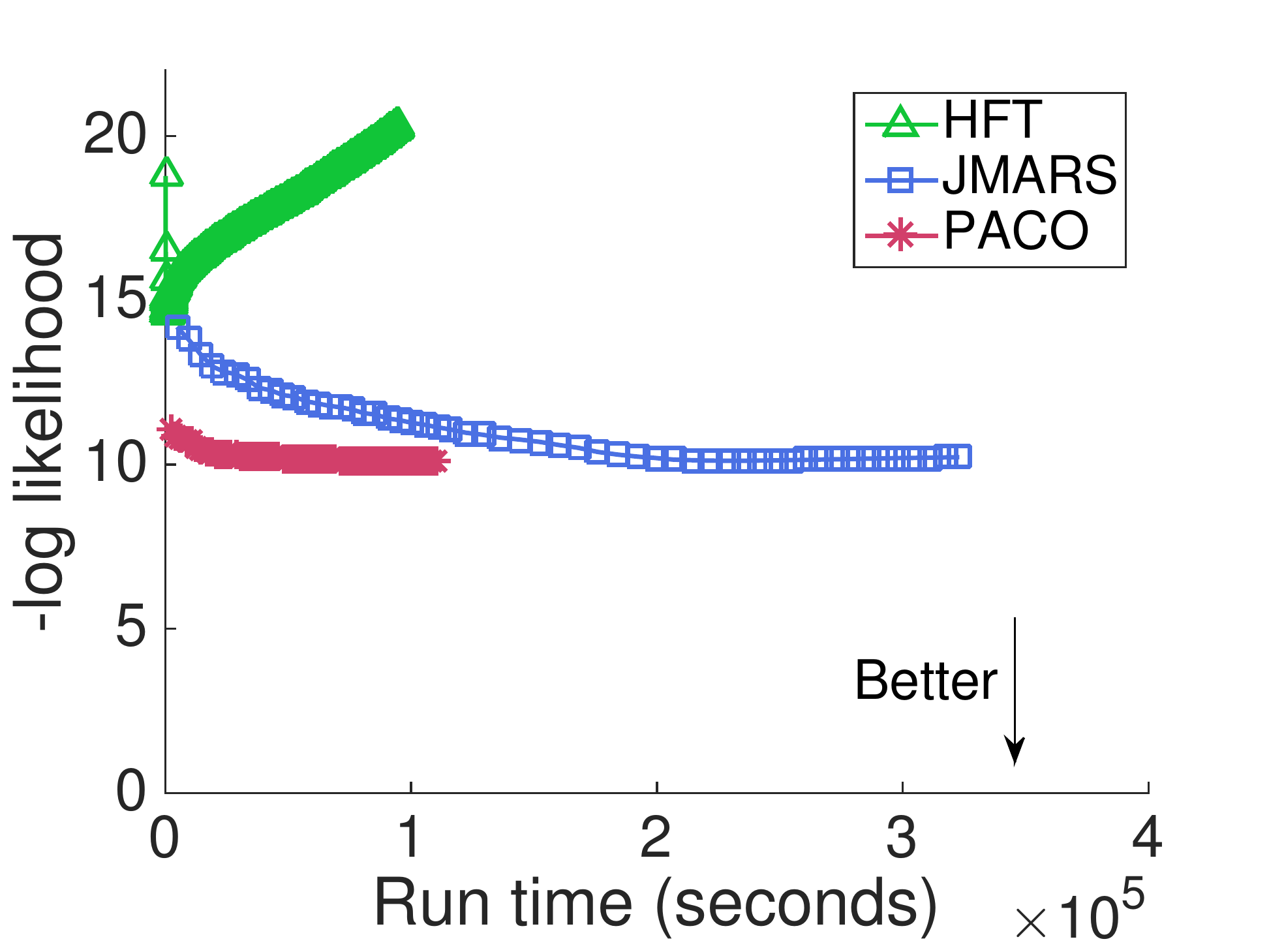}}
\subfigure[RateBeer]{\label{fig:joint_b}\includegraphics[width=0.32\textwidth]{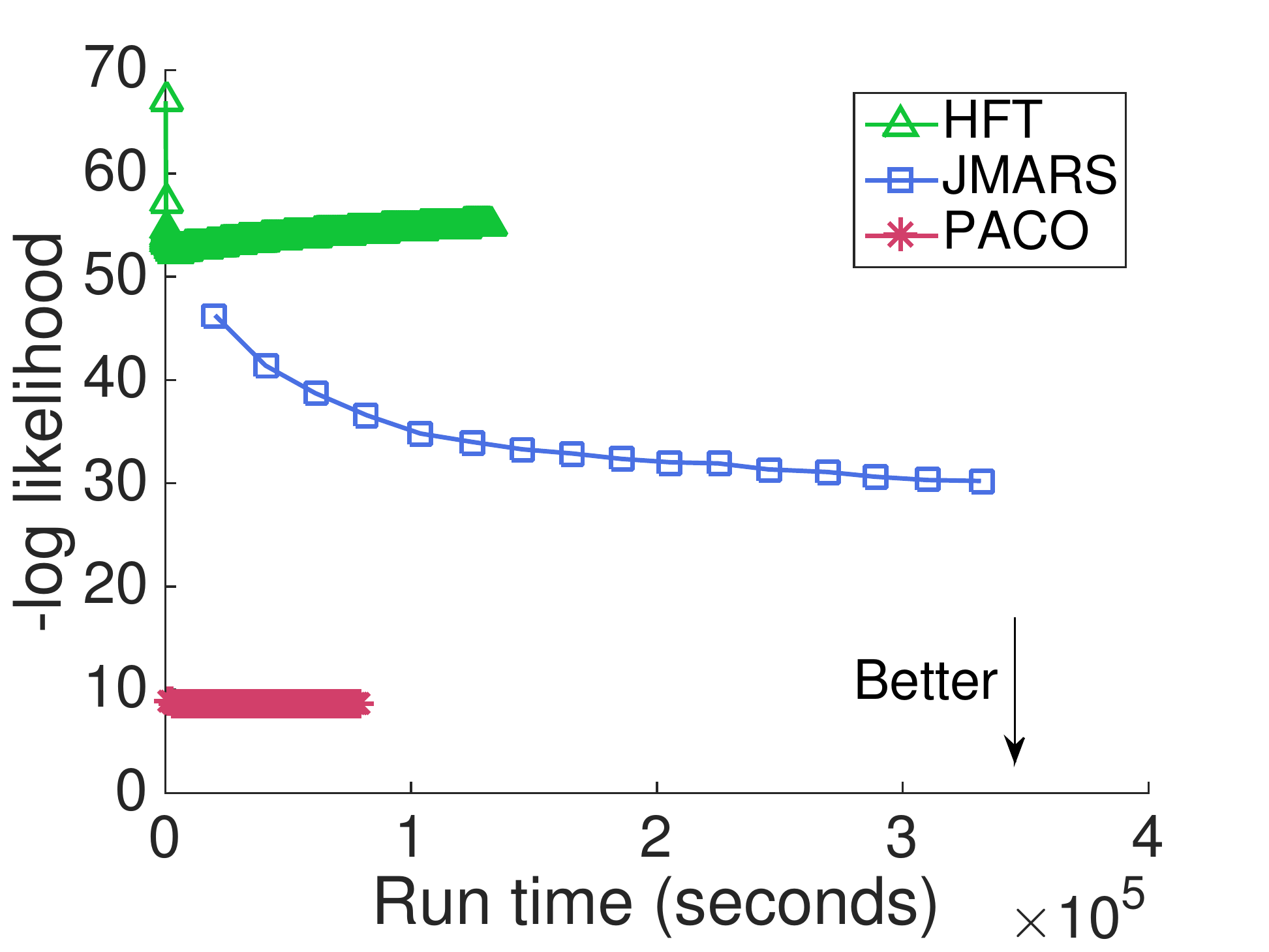}}
\subfigure[Yelp]{\label{fig:joint_c}\includegraphics[width=0.32\textwidth]{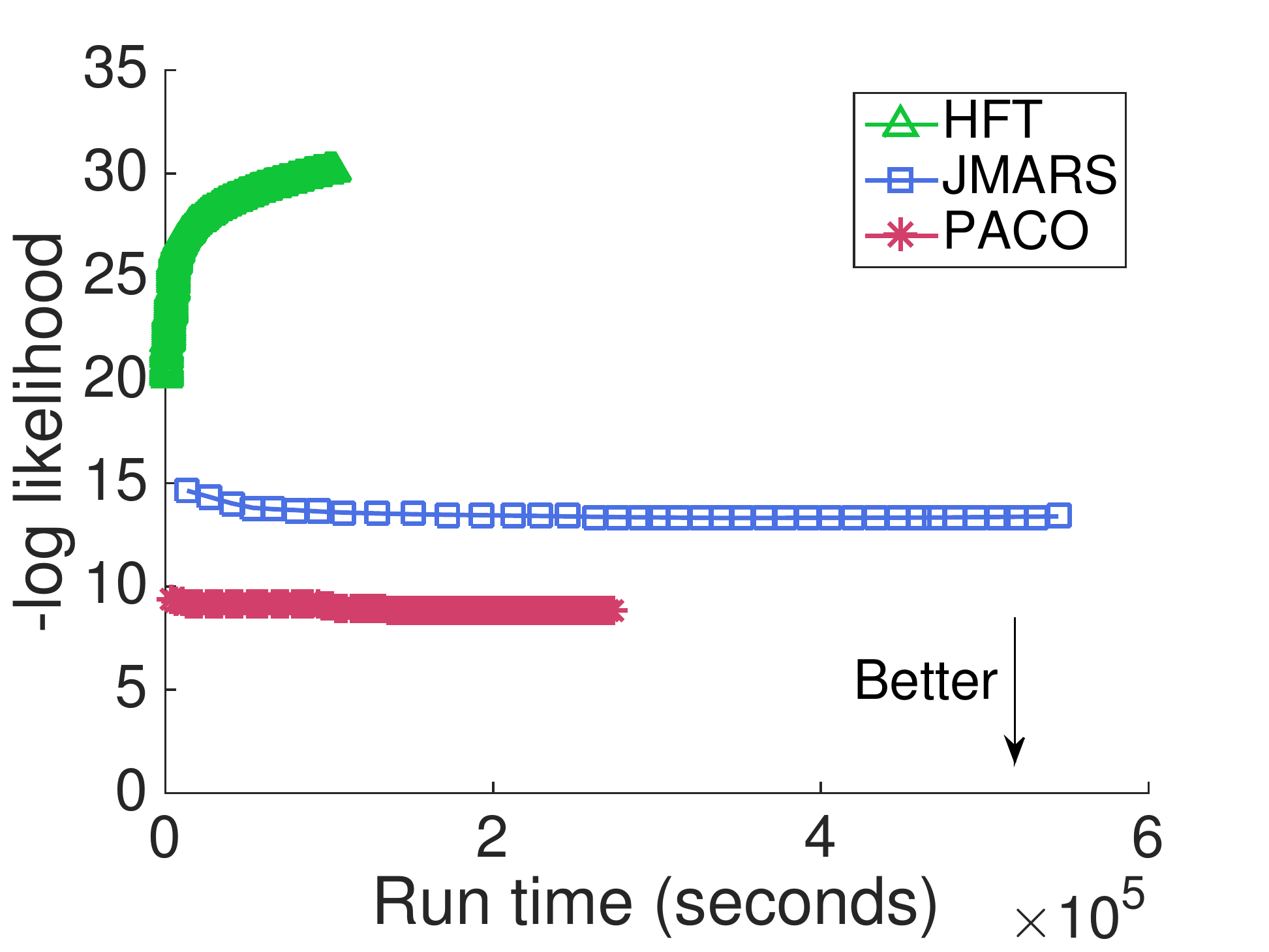}}
\caption{\textbf{Negative log likelihood:} \ourmodel better jointly predicts ratings and reviews than
state-of-the-art JMARS \cite{diao2014jointly} and HFT \cite{mcauleyhidden} on Amazon fine foods,
Yelp and RateBeer datasets.  The joint predictive power is captured by the
normalized negative log likelihood as described in 
\eqref{eq:joint_likelihood}.  Lower is better.} 
\label{fig:joint}
\end{figure}

With this approach we make a number of contributions:
\begin{itemize*}
	\item 
		We design a Poisson additive co-clustering model for backfitting word counts in documents.  
		We combine this with ACCAMS \cite{accams}
		to learn a joint Bayesian model of reviews and ratings, 
		with the ability to now interpret our model.
	\item We describe a new, efficient technique for sampling from a sum of Poisson
		random variables to facilitate efficient inference. It relies on
		treating discrete counts as ``residuals,'' similar to an additive
		regression model. 
	\item We give empirical evidence across multiple datasets that
		\ourmodel has better prediction accuracy for ratings than competing
		methods, such as HFT \cite{mcauleyhidden} and JMARS
		\cite{diao2014jointly}.  Additionally, our method predicts
		text reviews better than HFT, and achieves nearly as high quality review
		prediction as JMARS, while being far faster and simpler.
		As seen in Figure \ref{fig:joint}, \ourmodel outperforms both competing
		models in jointly predicting ratings and reviews.
\end{itemize*}
In summary, we propose a simple and novel model and sampler, capable of
characterizing user and item attributes very concisely, while
providing excellent accuracy and perplexity. 

\section{Related Work} 
\label{sec:Related Work}
Our model, while significantly different from previous approaches, touches upon
research from a variety of fields.

\paragraph{Collaborative Filtering} 
Collaborative filtering is a rich field of research. In particular, factorization
models that learn a bilinear model of latent factors have proven to be 
effective for recommendation problems
\cite{Koren2009MatrixFactorization,koren2010collaborative}.
A variety of papers have adapted frequentist intuition on factorization to 
Bayesian models
\cite{salakhutdinov2008Bayesian,mnih2007probabilistic,stern2009matchbox,CoBaFi}.
Of particular note are Probabilistic Matrix Factorization (PMF) and 
Bayesian Probabilistic Matrix Factorization (BPMF)
\cite{salakhutdinov2008Bayesian}, which we will compare to later.
Recently, ACCAMS \cite{accams} took a drastically different approach from
classic bilinear models. It uses an additive
model of co-clusterings to approximate matrices succinctly. While the resulting
model is simple and small, its prediction quality is as good as more complex
factorization models. 

A separate line of research in recommender systems has focused on using side
information to improve prediction quality.
This is particularly important when parts of the data are extremely sparse,
i.e. the cold start problem.
Content-based filtering is a popular approach to alleviate this
problem. Regression based latent factor models (RLFM)
\cite{agarwal2009regression} address cold-start problem by utilizing user and
item features. Cold-start users and items are able to share statistical
strength with other users and items through similarity in features space. fLDA
\cite{agarwal2010flda} uses text associated with items and user features to
regularize item and user factors, but does not make use of review text.

More recently, there has been a growing line of research on using Poisson
distributions in matrix factorization models
\cite{gopalan2014Bayesian,schein2015Bayesian,chaney2015probabilistic}.
This work shows the exciting potential uses of Poisson distributions for
understanding matrix data.  However, all of these models are left with limited
interpretability since they, too, rely on bilinear models.  Additionally, all of
these models rely on variational inference to learn the models.  Our work
provides the building blocks for using Poisson distributions in a wide array of
additive clustering applications and is the first work to learn a model of this
sort through Gibbs sampling rather than variational inference.

\paragraph{Review Mining and Modeling} 

Modeling online reviews has long been a focus of the data mining, machine
learning and natural language processing communities \cite{hu2004mining}.
Significant research has focused on understanding and finding patterns in
online reviews \cite{nocountry}.
More closely related to our work, a variety of papers model aspects and
sentiments of reviews \cite{lin2009joint,lazaridou2013Bayesian,jo2011aspect,kim2013hierarchical}.
For example, \cite{kim2013hierarchical} considers hierarchical structures in
aspects and sentiments. However, in these works ratings are not considered
jointly.  

\paragraph{Multimodal Models} 
Recently, there is increasing attention in jointly modeling review
text and ratings.  Collaborative topic regression (CTR)
\cite{wang2011collaborative} combines topic modeling and collaborative
filtering to recommend scientific articles.  HFT \cite{mcauleyhidden} jointly
models both ratings and reviews by designing link function to connect each
topic dimension to a latent factor, demonstrating improvements in
rating prediction.  RMR
\cite{ling2014ratings} relaxes the hard link between topic and latent factor
dimension for interpretable topics.  
\cite{xu2014collaborative} considers review texts with hidden user
communities and item groups relationship.  JMARS \cite{diao2014jointly} jointly models aspects,
sentiments, items, reviews, and ratings based on insights in review structure.

A related line of work models multi-aspect ratings \cite{mcauley2012learning,titov2008joint}.
However, these works often rely on availability of aspect-specific ratings,
which are often not available. In contrast, our models learns sentiments in
different aspects without requiring multi-aspect ratings. 

\paragraph{Additive Clustering}
Our model, built on additive co-clustering, takes a
different approach from classic collaborative filtering literature.
Our approach is conceptually similar to older literature from
psychology on additive clustering
\cite{shepard1979additive,tenenbaum1996learning,navarro2008latent}.
ADCLUS \cite{shepard1979additive} argues that similarity
between objects should be based on similarity on a subset of discrete
properties.  This perspective reinforces our belief that additive
co-clustering is a good fit for user behavior modeling.

From a computational perspective, ACCAMS \cite{accams}, described above, focuses on an additive
\emph{co-clustering} model of Gaussian distributions and describes a collapsed
sampler for efficient learning.  
\cite{palla2012infinite} uses a sum of clusters within a logistic function for
modeling binary data but has difficulty scaling.  
Here we demonstrate that we can successfully scale learning a sum of clusters within a Poisson distribution.

\begin{table}
	\centering
	\begin{tabular}{l|l}
		\toprule
		Symbol & Definition \\ 
		\midrule
		$N,M$ & Number of rows (users) and columns (items) \\ 
		$\data$ & Data matrix $\in \mathbb{R}^{N \times M}$ (with missing values) \\ 
		$\Ical$ & Indicator matrix $\in \{0,1\}^{N \times M}$ for $\data$ \\ 
		$\nstencils$ & Number of stencils \\ 
		$k_n^\sindexp,k_m^\sindexp$ & Number of user and item clusters in stencil $\sindex$ \\ 
		$\template^\sindexp$ & Matrix $\in \mathbb{R}^{k_n^{\sindexp} \times k_m^{\sindexp}}$ for stencil $\sindex$ \\ 
		$\uclusters^\sindexp$ & Vector of user assignments $\in \{1,\ldots,k_n^\sindexp \}^N$ \\ 
		$\mclusters^\sindexp$ & Vector of item assignments $\in \{1,\ldots,k_m^\sindexp \}^M$ \\ 
		$\Sfn(\template,\uclusters,\mclusters)$ & $\in \mathbb{R}^{N \times M}$ defined by $\Sfn(\template,\uclusters,\mclusters)_{u,m} =\template_{\uclusters_u,\mclusters_m}$ \\ 
		$\mathcal{W}$ & Set of all words used in the reviews \\ 
		$n_{u,m,x}$ & Count for word $x$ in review $(u,m)$ \\ 
		$\mu^{(i)}_x$ & Rate of Poisson in language model $i$ for word $x$ \\ 
		$\mathcal{L}_{u,m}$ & Set of language models used for review $(u,m)$ \\ 
		\bottomrule
	\end{tabular}
	\vspace{-3mm}
	\caption{Symbols used throughout this paper.}
	\label{tab:notation}
\end{table}

\section{Additive Co-Clustering}

We begin by giving a general overview of additive co-clustering in
order to gain familiarity with the notation and problem.
A full list of the symbols used can be found in Table \ref{tab:notation}.
We will use bold uppercase letters to denote matrices, bold lower case letters
to denote vectors, and non-bold letters for scalars.
To index into a matrix or vector we will use subscripts, e.g. $\data_{u,m}$
refers to the value in row $u$ and column $m$ of $\data$; we will use ``:'' to
select an entire row or column in a matrix when necessary.
Superscripts are often used to denote different instances of the object.

Our goal is to jointly model documents and review scores.  For the
sake of completeness, we briefly review the formulation of additive
co-clustering for Gaussian data, as described in ACCAMS.  We focus on
the problem of matrix approximation, where we have a real valued
(often sparse) matrix, where we observe a small percentage of the
entries in the matrix and our goal is to predict the missing values.
To be precise, we have a matrix $\data \in \RR^{N \times M}$ with the
set $\Ical$ of observed entries $(u,m)$.  To predict the missing
values, we would like to learn a model $\mathcal{M}$ with parameters
$\theta$, such that the size of $\theta$ is small,
$|\theta| \ll |\Ical|$, and $\mathcal{M}(\theta)$ approximates $\data$
well:
\begin{align}
	\mini_\theta \sum_{(u,m) \in \Ical} (\data_{u,m} - \mathcal{M}(\theta)_{u,m})^2
\end{align}                              
The concept of a small model in the context of behavior modeling has
typically been captured by low rank factorization of the rating matrix.  We
consider the generalization of this concept, defining a small model by the
number of bits required to store it.

Key to our model is the notion of a stencil, an extremely easy way to
represent a block-wise constant rank-$k$ matrix.  A stencil $\template$ assigns each
row $u$, typically a user, to a row cluster $\clindex$, each column, typically an item, to a column cluster
$\clindexm$. The rating that user $u$ gives to item $m$ is thus predicted by the value $\template_{\clindex,\clindexm}$. Formally:
\begin{definition}[Stencil]
  A stencil $\Sfnp$ is a block-wise constant matrix $\Sfn \in \RR^{N
    \times M}$ with the 
  property that 
  $$\Sfnp_{u,m} = \template_{\uclusters_u, \mclusters_m}
  \text{ for a template }
  \template \in \RR^{k_n \times k_m}$$ 
  and index vectors
  $\uclusters \in \cbr{1, \ldots, k_n}^N$ and \smash{$\mclusters \in \cbr{1,
  \ldots, k_m}^M$}.
\end{definition}
Therefore, our goal is to find a
stencil $\Sfnp$ with a small approximation error 
$\data - \Sfnp$.

We observe that storing a stencil with small $k_n,k_m$ is quite compact, with
the cost in bits bounded by \cite{accams}:
\begin{align}
  \mathrm{Bits}(\{\template,\uclusters,\mclusters\}) = N\log_2k_n + M\log_2k_m + 32k_nk_m
  \label{eq:stencil_cost}
\end{align}
Note that this bound is very loose, as it ignores any properties of
the distribution of $\uclusters$ and $\mclusters$. That said, even the
worst case bound is much tighter than what can be accomplished with
inner product models. To improve the approximation accuracy without
making the model size grow quickly, we can use \emph{multiple} stencils
in an additive fashion:
\begin{align*}
  \mini_{\cbr{\template^\sindexp, \uclusters^\sindexp,
  \mclusters^\sindexp}} \sum_{(u,m) \in \Ical} \left(\data_{u,m} - \sum_{\sindex=1}^\nstencils \Sfnpi_{u,m}\right)^2
\end{align*}
That is, we would like to find an additive model of $\nstencils$ stencils that minimizes
the approximation error to $\data$. 

Unfortunately, finding linear combinations of co-clusterings is NP-hard.  It is
easy to see this by reducing co-clustering, which is NP-hard, to our problem by
setting $\nstencils = 1$. In \cite{accams}, we showed how to solve this problem using a back fitting algorithm. 
To make the problem tractable, we assumed that user and item clusters in each stencil follow a Chinese restaurant process (CRP), 
and that the observed ratings are normally distributed across the mean of each co-cluster. Formally speaking, we have the following generative model  \cite{accams}:

\begin{align}
  \label{eq:statmodel}
  \data_{u,m} & \sim \Ncal\rbr{\sum_{\sindex=1}^\nstencils
                \Sfnpi_{u,m}, \sigma^2} \\
  \uclusters^\sindexp, \mclusters^\sindexp & \sim \mathrm{CRP}(\delta)
                                             \text{ for all } \sindexp \\
  \template^\sindexp_{c,d} & \sim \Ncal(0, \sigma^2_\sindexp)                                
\end{align}

The benefit of this model is that conditioned on all stencils but one,
the problem of inferring $\template^\sindexp$ becomes one of inferring
Gaussian random variables for the rating (i.e. $\template^\sindexp_{c,d}$ ). Likewise, inferring $\uclusters^\sindexp$
and $\mclusters^\sindexp$ is a straightforward clustering problem. Thus, the backfitting  algorithm sweeps through all stencils, one at a time, estimating the stencil's users and items cluster assignments in addition to  the co-clusters ratings.  We have shown in  \cite{accams}, that fitting  stencil $\ell$ while fixing other stencils, is equivalent to fitting the residual error  between the observed ratings and the current  estimate using all  stencils but stencil $\ell$. This algorithm converges with  guarantees as shown in  \cite{accams}.

\section{Poisson Additive Co-Clustering} 
\label{sec:Model}

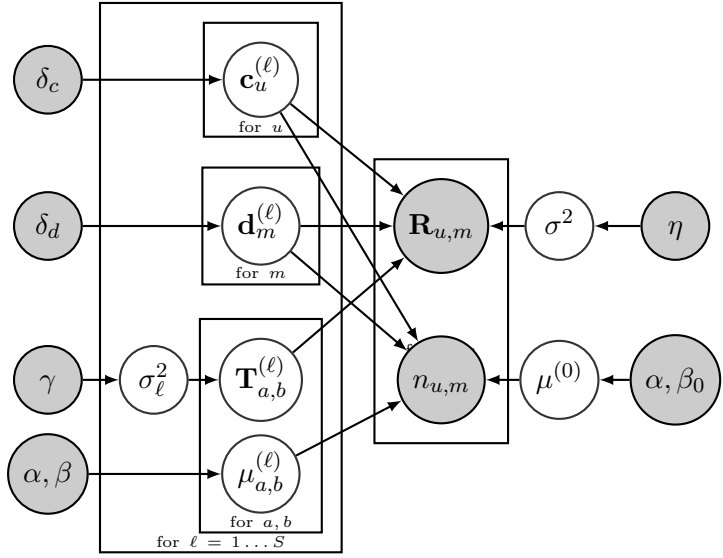
\begin{figure}[bt]
	\centering
\begin{tikzpicture}[>=latex,text height=1.5ex,text depth=0.25ex]
  \matrix[row sep=0.8cm,column sep=0.4cm] {
    \node (delta_c) [observed]{$\delta_c$}; & &
    \node (c) [latent]{$\uclusters_u^\sindexp$}; 
	\\
    \node (delta_d) [observed]{$\delta_d$}; & & 
    \node (d) [latent]{$\mclusters_m^\sindexp$}; & & &
    \node (r) [observed]{$\data_{u,m}$}; &
    \node (sigma) [latent]{$\sigma^2$}; &
    \node (gamma) [observed]{$\eta$}; 
    \\
    \node (gammatau) [observed]{$\gamma$}; &
    \node (tau) [latent]{$\sigma_\sindex^2$}; &
    \node (S) [latent]{$\template_{\clindex,\clindexm}^\sindexp$}; & & &
    \node (textdata) [observed]{$\textdata_{u,m}$}; &
	\node (baselang) [latent]{$\mu^{(0)}$}; &
	\node (baselangprior) [observed]{$\alpha,\beta_0$}; &
    \\[-7mm]
	\node (blocklangprior) [observed]{$\alpha,\beta$}; & &
	\node (blocklang) [latent]{$\mu^\sindexp_{\clindex,\clindexm}$}; &
	\\
   };
  \path[->]
  (gammatau) edge[thick] (tau)
  (tau) edge[thick] (S)
  (delta_c) edge[thick] (c)
  (delta_d) edge[thick] (d)
  (c) edge[thick] (r)
  (d) edge[thick] (r)
  (S) edge[thick] (r)
  (sigma) edge[thick] (r)
  (gamma) edge[thick] (sigma)
  (c) edge[thick] (textdata)
  (d) edge[thick] (textdata)
  (blocklang) edge[thick] (textdata)
  (blocklangprior) edge[thick] (blocklang)
  (baselang) edge[thick] (textdata)
  (baselangprior) edge[thick] (baselang)
  ;
  \begin{pgfonlayer}{background}
    \node (usercluster) [plate, fit=(c)] {\
      \\[9mm]\tiny for $u$};
    \node (itemcluster) [plate, fit=(d)] {\
      \\[9mm]\tiny for $m$};
    \node (stencil) [plate, fit=(S) (blocklang)] {\
      \\[22.5mm]\tiny for $\clindex,\clindexm$};
    \node (ratings) [plate, fit=(r) (textdata)] {\
      \\[10mm]\tiny for $u,m$};
    \node (stencils) [plate, fit=(usercluster) (itemcluster) (tau) (stencil)] {\ 
      \\[67mm]\tiny for $\sindex = 1 \ldots \nstencils$};
    \end{pgfonlayer}
\end{tikzpicture}
\caption{The generative model for \ourmodel to predict both ratings $\data$ and
review text $\textdata$.  (Note, for the sake of space we simplify the model
slightly by not explicitly separating the different language models associated
with each stencil.)
}
\label{fig:model}
\end{figure}

While ACCAMS \cite{accams} explains the ratings matrix well, it does not utilize
the reviews associated with those ratings.  In this section we introduce PACO,
a Poisson Additive co-clustering model that jointly model text and reviews.
PACO builds on the idea of stencils introduced in ACCAMS. Each stencil
$\template$ assigns each user $u$ to a cluster say $a$ and each item $m$ to a
cluster say $b$. Given the block (i.e. co-cluster) denoted by $\template_{a,b}$,
we design a model to jointly generates both the rating user $u$ gives to item
$m$ as well as the review she wrote for this item. We endow each block with a
Gaussian distribution that model the mean rating associated with cluster $a$
and $b$. The question now is: how can we parametrize the text model of block
$(a,b)$?

\subsection{Modeling Reviews using an Additive Poisson Model}

A standard approach in the text mining literature is to model reviews
using a multinomial distribution; however, in PACO, as in ACCAMS, we
want to combine multiple stencils to enhance the model. While it is
easy to define an additive model over review scores, it is nontrivial
to accomplish this using multinomial distributions for reviews.  Quite
obviously, if $p$ and $q$ are multinomial probability distributions,
then $p + q$ is no longer in the probability simplex. Instead, the
transform is given by $(p+q)/\nbr{p+q}_1$, thus making updates highly
nontrivial, since additivity of the model is lost, which means that we
would need to update the \emph{entire language model} whenever even
just a single stencil changes.  This would make a backfitting
algorithm very expensive.

Rather, we introduce a novel approach to modeling using the Poisson
distribution. In a nutshell, we exploit the fact that the Poisson distribution  is closed
under addition, i.e.\ for
\begin{align}
  \nonumber
  a \sim \mathrm{Poi}(\lambda) \text{ and } 
  b \sim \mathrm{Poi}(\gamma) \text{ we have }
  a + b \sim \mathrm{Poi}(\lambda + \gamma)
\end{align}
where $\lambda$ and $\gamma$ denote the rates of each of the random
variables, i.e.\ $\Eb[a] = \lambda$ and $\Eb[b] = \gamma$. 

For each user and item pair $(u,m)$ pair we let $n_{u,m,x}$ denote the count
 for word $x$ in the review. We now design an
additive model for each review. The idea is that the distribution over
each word is given by a sum over Poisson random variables with rates
$\mu_{\uclusters_u^{(\ell)},\mclusters_m^{(\ell)},x}^{(\ell)}$ for each word $x$ across all stencils. The benefit of this
approach is that we no longer need to ensure normalization.  We will detail the generative model in the following subsection. 

\subsection{The Joint Generative Model}

Now we are ready to present the full model. To design a joint model we face an
important challenge: we need to assess whether to perform good
recommendation or whether we strive to optimize for good
perplexity. In the former case, it is undesirable if the reviews carry
the majority of the statistical weight. Hence it is worthwhile to
normalize the reviews by their length. This technique is common in NLP literature
\cite{wang2005group,wang2006topics}.  This yields the following joint objective:

\begin{align*}
  \mini_{\cbr{\template^\sindexp, \uclusters^\sindexp,
  \mclusters^\sindexp, \lambda}} &\sum_{(u,m) \in \Ical} \left(\data_{u,m} - \sum_{\sindex=1}^\nstencils \Sfnpi_{u,m}\right)^2
  \\ &+ \sum_{(u,m) \in \Ical} \frac{1}{|n_{u,m}|_1}\sum_{x \in \mathcal{W}} \log \mathrm{Poi}( \lambda_{u,m,x})
\end{align*}
where
\begin{align}
	\lambda_{u,m} &= \mu^{(0)} \!+\! \mu^{(m)} \!+\! \left[ \sum_{\ell=1}^S \mu_{\uclusters_u^{(\ell)}, \mclusters_m^{(\ell)}}^{(\ell)} \!+\! \mu_{\mclusters_m^{(\ell)}}^{(m,\ell)} \!+\! \mu_{\uclusters_u^{(\ell)}}^{(u,\ell)} \right] \label{eq:lambda_review}
\end{align}

Our goal is to learn a set of stencils whose summation minimizes the prediction error on ratings and maximizes the likelihood of generating the text. 
To model review text, we allow each stencil to have three language models: a stencil-specific user language model $\mu_{\uclusters_u^{(\ell)}}^{(\ell)}$, a stencil-specific item language model $\mu_{\mclusters_m^{(\ell)}}^{(m,\ell)}$,  and block language model, $\mu_{\uclusters_u^{(\ell)}, \mclusters_m^{(\ell)}}^{(\ell)}$. The block language model captures the stencil-specific interaction between the item and the user.  In addition, we add a global item language model, $\mu^{(m)}$, and a global background language model,  $\mu^{(0)}$.  The text of the review is modeled as a combination of these Poisson language models.

Minimizing the aforementioned objective function is approximately equivalent to maximizing the log-likelihood of the graphical model in Figure \ref{fig:model}. The generative process proceeds as follows:
\begin{align}
	\data_{u,m} & \sim \Ncal\rbr{\sum_{\sindex=1}^\nstencils
	\Sfnpi_{u,m}, \sigma^2} \\
	\uclusters^\sindexp, \mclusters^\sindexp & \sim \mathrm{CRP}(\delta)
	\text{ for all } \sindexp \\
	\template^\sindexp_{c,d} & \sim \Ncal(0, \sigma^2_\sindexp) \\
	n_{u,m,x} &\sim {\mathrm{Poi}}\rbr{\lambda_{u,m,x}} \label{eq:poisson} \\
	\mu^{(*)}_x &\sim {\rm Gamma}(\alpha,\beta) \label{eq:poisson_priors}
\end{align}
where $\uclusters_u^\sindexp, \mclusters_m^\sindexp$ is the cluster $(u,m)$
assigned to in stencil $\sindex$.  In essence, we model user and item clusters inside each stencil using a Chinese restaurant process (CRP). Ratings are modeled similar to ACCAMS while \eqref{eq:lambda_review}, \eqref{eq:poisson} and
\eqref{eq:poisson_priors} contain the additional text modeling aspects of our model.
To avoid overfitting, we use conjugate Gamma prior for all the vectors $\mu$: 
\begin{align*}
	P(\mu_x) = \frac{\beta^\alpha}{\Gamma(\alpha)}\mu_x^{\alpha - 1} e^{-\beta\mu_x} 
\end{align*}
Since the mode of the Gamma distribution is $\frac{\alpha - 1}{\beta}$, a very large
$\beta$ should ensure that only a small amount of data is blamed. 
Given this model, we now consider the challenge of learning the parameters in practice.

\section{The Sampling Algorithm}
\label{sub:sampler}

The goal of inference is to learn a posterior distribution over stencils'
parameters which are: user and item cluster assignments, stencil ratings of
each block, and the multiple language models.
To do this, we use Gibbs sampling.

Jointly sampling text and rating adds significant complications over just sampling ratings data and requires a novel sampling technique.   In particular, our sampler offers (1) a new novel technique to
learn the sum of Poisson rates $\mu$ and (2) an efficient method for sampling
cluster assignments based on both the text model and ratings model.
We describe each of these challenges and solutions below in Section \ref{sec:sample_poisson}, 
followed by the complete learner shown in Algorithm
\ref{alg:full}, which combines the new, novel Poisson sampler and ACCAMS's sampler for
Gaussian rating data.

\subsection{Sampling a Sum of Poisson Distributions} \label{sec:sample_poisson}
The Gamma distribution is conjugate to the Poisson distribution, typically
allowing for an easy sampling of the Poisson's rate $\lambda$ from the Gamma
distribution.  However, in this case we have a sum of Poisson distributions and 
we would like to sample the rate of each of these distributions.  

To make this tractable, we create a multinomial the from rates of the involved Poisson distributions
and sample form this multinomial the fraction of counts coming from each Poisson distributions.
To be precise, for a particular
$n_{u,m,x}$ we have $\lambda_{u,m,x} = \sum_{i \in \mathcal{L}_{u,m}}
\mu^{(i)}_{x}$, where $\mathcal{L}_{u,m}$ is the set of Poisson distributions
from which words in $(u,m)$  are sampled.  We define:
\begin{align}
	\cbr{\mu^{(*)}_{u,m,x}} := \cbr{\mu_{x}^{(i)}}_{\forall i \in \mathcal{L}_{u,m}}
  \text{ and }
  \cbr{\hat{n}_{u,m,x}} := \cbr{\hat{n}_{u,m,x}^{(i)}}_{\forall i \in \mathcal{L}_{u,m}}
\end{align}
We can therefore sample $\cbr{\hat{n}_{u,m,x}}$ by
\begin{align}
	\cbr{\hat{n}_{u,m,x}} &\sim {\rm Multi}\rbr{\frac{\cbr{\mu^{(*)}_{u,m,x}}}{\lambda_{u,m,x}}, n_{u,m,x}}. \label{eq:n_sampler}
\end{align}
The result is
$n_{u,m,x} = \sum_{i \in \mathcal{L}_{u,m}} \hat{n}_{x}^{(i)}$.
 That is, if we observe $n_{u,m,x}$ occurrences of word $x$ in review $(u, m)$, we break this count to a set of $ \hat{n}_{x}^{(i)}$ counts, each of which are credited to the 
 corresponding Poisson distribution $i$, where $i$ indexes over the set of involved Poisson distributions.  For example, if the word ``delicious'' is used three
times in a review, we may consider one use of the word to be ``from'' the base
language model and two uses of the word to be ``from'' the item-specific
language model.

By sampling these count allocations, we can now tractably sample our Poisson
rates and later our cluster assignments.
To sample a particular $\mu^{(i)}_x$, we consider $\mathcal{R}_i$, the set of
reviews $(u,m)$ partially sampled form a Poisson distribution with rate $\mu^{(i)}$.
Therefore, we can sample $\mu^{(i)}_x$ by:
\begin{align}
	\mu_{x}^{(i)} &\sim \Gamma\rbr{\alpha^{(i)}, \beta^{(i)}} \!\prod_{(u,m) \in \mathcal{R}_i} \!{\mathrm{Poi}}
	\rbr{\left. \hat{n}_{u,m,x}^{(i)}  \right| \mu_{x}^{(i)}} 
	\\  &\sim \Gamma\!\rbr{\alpha^{(i)} \!+\!\! \sum_{(u,m)\in \mathcal{R}_i}\! \hat{n}_{u,m,x}^{(i)}, \beta^{(i)} + \abr{\mathcal{R}_i} } \label{eq:mu_sampler}
\end{align}
This is trivially parallelized across all words in a particular language model
$\mu$, which we will expand on later.

\subsection{Sampling Cluster Assignments}  \label{sec:sample_clusters}
For each stencil, we need to sample the cluster assignment for each user and item. 
To do this for users, we need to calculate the posterior distribution $p(\uclusters_u^\sindexp = \clindex | {\rm rest}
)$ for each cluster $\clindex$.  This probability is composed of three main terms: (1)
the CRP prior, (2)
the likelihood of the user ratings and (3) the likelihood of the user review texts.  This can be written as
\begin{align*}
	p(\uclusters_u^\sindexp = \clindex | {\rm rest}) \propto {\rm CRP}(\clindex) 
	\prod_{(u,m) \in \mathcal{R}_u} p(\mathbf{R}_{u,m} | {\rm rest})
	\prod_x p(n_{u,m,x}  | {\rm rest})
\end{align*}
where $\mathcal{R}_u$ denotes the set of reviews from user $u$. 
This probability in log-space (incorporating review normalization) becomes:
\begin{align}
	\log p(\uclusters_u^\sindexp = \clindex | {\rm rest}) \propto \log \left({\rm CRP}(\clindex)\right) \label{eq:sample_cluster}
  + \hspace{-2mm} \sum_{(u,m) \in \mathcal{R}_u} \log p(\mathbf{R}_{u,m} | {\rm rest})
	+ \frac{1}{|n_{u,m}|}\sum_x \log p(n_{u,m,x}  | {\rm rest})
\end{align}
The details for calculating  the CRP term and rating terms can be
found in \cite{accams}.  Here, we focus on how to efficiently calculate the term that corresponds to
probability of the reviews.

To calculate the probability of the text, we in fact focus on the
$\hat{n}_{u,m,x}$ rather than $n_{u,m,x}$.  Specifically, when sampling the user cluster assignment, we must calculate on
\begin{align}
  \Delta_{u,\clindex} &:= \sum_{ (u,m) \in \mathcal{R}_u} \sum_x 
	\log\;{\mathrm{Poi}}\rbr{\left. \hat{n}^\sindexp_{u,m,x} \right| \mu^\sindexp_{\clindex,\mclusters^\sindexp_m,x} } \nonumber
  + \log\;{\mathrm{Poi}}\rbr{\left. \hat{n}^{(u,\sindex)}_{u,m,x} \right| \mu^{(u,\sindex)}_{\clindex,x} } \nonumber
	\\ &= \sum_{(u,m) \in \mathcal{R}_u}\sum_x
             \hat{n}^\sindexp_{u,m,x}\log \sbr{\mu^\sindexp_{\clindex,\mclusters^\sindexp_m,x}} - \mu^\sindexp_{\clindex,\mclusters^\sindexp_m,x} 
  + \hat{n}^{(u,\sindex)}_{u,m,x}\log \sbr{\mu^{(u,\sindex)}_{\clindex,x}} - \mu^{(u,\sindex)}_{\clindex,x} \label{eq:logdelta}
\end{align}
For $k$ clusters, to naively calculate $\Delta_{u,\clindex}$ for each
user would require $O(k|\mathcal{W}|)$ logarithm evaluations, which
are significantly slower than the other simple addition and
multiplication operations.  However, we can accelerate this
significantly. Define the following terms:
\begin{align*}
  \tilde{\mu}_{\clindex,\clindexm}^\sindexp & = 
  \sum_x \mu_{\clindex,\clindexm,x}^\sindexp
                                              && \text{ and } & 
   \tilde{\mu}_{\clindex}^{(u,\sindex)} & = \sum_x \mu_{\clindex,x}^\sindexp 
  \\
	\eta_{u,\clindexm} & = \sum_{\substack{m | \mclusters^\sindexp_m = \clindexm, \\ (u,m) \in \mathcal{R}_u}} 1
   && \text{ and } &
   \hat{\eta}^\sindexp_{u,\clindexm} & = \sum_{\substack{m | \mclusters^\sindexp_m = \clindexm, \\ (u,m) \in \mathcal{R}_u}} \hat{n}^\sindexp_{u,m} 
   && \text{ and } &
  \hat{\eta}^{(u,\sindex)}_u & = \sum_{(u,m) \in \mathcal{R}_u} \hat{n}^{(u,\sindex)}_{u,m}
\end{align*}
All of these terms can be pre-calculated and cached for sampling all
cluster assignments for stencil $\sindex$.
Now we can rearrange the terms of \eqref{eq:logdelta} to achieve the following simplified equation:
\begin{align}
  \label{eq:sample_cluster_condensed}  \Delta_{u,\clindex} &= - \sum_{\clindexm = 1}^k \eta_{u,\clindexm} 
	\left(\tilde{\mu}_{\clindex,\clindexm}^\sindexp + \tilde{\mu}_{\clindex}^{(u,\sindex)} \right)
	+ \inner{ \hat{\eta}^\sindexp_{u,\clindexm}}{\log \mu^\sindexp_{\clindex,\clindexm}}
	+ \inner{\hat{\eta}^{(u,\sindex)}_{u}}{\log \mu^{(u,\sindex)}_{\clindex}} 
\end{align}
With this formulation we can cache the logarithm of each
$\mu_{\clindex,\clindexm,x}$ and $\mu_{\clindex,x}$ and reuse them for sampling cluster assignment of each user. 
As such, we only need to take one pass over the original data and thus  minimize
the number of logarithm computations in each iteration. 
Sampling item cluster assignments can be done by an analogous sampler.
We will show in Section \ref{sec:implementation} that this approach results in a fast sampling procedure.

Now we are ready to describe our full Gibbs sampling algorithm. For each stencil, we first update the rating model (see \cite{accams} for details) and then update the text model using aforementioned techniques: 
Specifically, we resample $\hat{n}$ following \eqref{eq:n_sampler}, and then resample the stencil's $\mu$ terms (i.e. the Poisson language models associated with that stencil) 
based on $\hat{n}$, following \eqref{eq:mu_sampler}.
Finally, we update the cluster assignments for a given stencil following \eqref{eq:sample_cluster}. 
The full procedure is given in Algorithm \ref{alg:full}.

\begin{algorithm}  
\caption{\ourmodel Sampler}
\label{alg:full}                       
\begin{algorithmic}           
    \STATE {Run K-Means ACCAMS for initial cluster assignments.} 
	\STATE {Initialize all $\mu^{(i)}$ to a vector $\mathbf{1}$}
    \WHILE {not converged }
	\STATE {Resample  $\{\hat{n}_{u,m,x}\}$ for all $u,m,x$ by \eqref{eq:n_sampler}}
    \STATE {Sample $\mu^{(0)}$ by \eqref{eq:mu_sampler}}
    \FOR {Stencil $\sindex = 1, \dots, S$}
      \STATE {Update predicted ratings as described in \cite{accams}}
      \IF {Not first iteration}
		\STATE {Resample $\{\hat{n}_{u,m,x}\}$ for all $u,m,x$ by \eqref{eq:n_sampler}}
      \ENDIF
	\STATE {Sample $\mu^\sindexp, \mu^{(m,\ell)}, \mu^{(u,\ell)}$ by \eqref{eq:mu_sampler}}
	\STATE {Sample $\uclusters_u^\sindexp$ by \eqref{eq:sample_cluster}}
	\STATE {Sample $\mclusters_m^\sindexp$ analogous to \eqref{eq:sample_cluster}}
    \ENDFOR
    \STATE {Sample $\mu^{(m)}$ by \eqref{eq:mu_sampler}}
    \ENDWHILE
\end{algorithmic}
\end{algorithm}

\subsection{Implementation}
\label{sec:implementation}

The sampler is written in C++11. The implementation is very efficient and  uses  the following techniques. 
Sampling $\cbr{\hat{n}_{u,m,x}}$ is embarrassingly parallelizable across reviews. 
With $\mu$ fixed, this involves simply parallel sampling from the re-parametrized multinomials\footnote{GNU Scientific Library (GSL) is used for multinomial sampling. OpenMP is used to parallelize independent sampling.}.
Sampling the Poisson rate $\mu$ is also embarrassingly parallelizable across $\mu$ and across words. 
We use C++11 implementation of Gamma sampler. 
The review likelihood can be efficiently calculated when sampling user/item cluster assignments. 
With $\tilde\mu_{a,b}^{(\ell)}$, $\tilde\mu_{a}^{(u,\ell)}$, $\log(\mu_{a,b,x}^{(\ell)})$ and $\log(\mu_{a,x}^{(u,\ell)})$ cached, 
the review likelihood can be calculated with one pass over non-zero words. Again, this can be parallelized across users/items. 

\textbf{Speed:}
We generally found the above optimizations to make our sampler sufficiently
fast for the large datasets we tested on.
On the Amazon fine foods dataset described below, sampling all cluster
assignments for ons stencil takes 3.66 seconds on average, sampling all
$\hat{n}$ for half million reviews takes 45 seconds, and sampling all $\mu$ for one stencil takes less
than one second on a single machine.

\section{Experiments}
We now test our model in a variety of settings both to understand how it models
different types of data and to demonstrate its performance against similar,
recent models.

\subsection{Experimental Setup} 

\paragraph{Datasets} 
To test our model, we select four datasets about movies, beer,
businesses, and food.  All four datasets come from different websites and communities, thus
capturing different styles and patterns of online ratings and reviews.
In all of these datasets, one observed (user,item) pair is associated with one rating and
one review. 
We randomly select 80\% of data as training set and 20\% as testing set while
making sure every user and item in the testing set has at least one example in training
set. Infrequent words, standard stop words and words shorter than 3 characters
are removed. We rescale ratings to the same range during training and center
all ratings based on the global average in the dataset. 
The resulting datasets are summarized in Table \ref{tab:data}.

\begin{table}[tbh]
\small
\centering
 \begin{tabular}{lrrrr}
 \toprule
 Dataset & Yelp & Food &RateBeer & IMDb\\
  \midrule
 \# items &60,785&74,257&110,369& 117,240\\
\# users &366,715&256,055&29,265& 452,627\\
\# observations & 1,569,264 & 568,447 & 2,924,163& 1,462,124\\
\# unigrams  & 9,055 & 9,088 & 8,962& 9,182\\
avg. review length  & 45.20 & 31.55 & 28.57 & 88.30\\
   \bottomrule
 \end{tabular}
 \caption{Four datasets used in experiments. 
 Infrequent words, standard stop words and words shorter
 than 3 characters are removed during pruning.}\label{tab:data}
 \end{table}
 
\paragraph{Metrics}
To evaluate our review model, we examine its ability to predict held-out
testing ratings and reviews. 

{\bf RMSE:}
When comparing predictions of held-out ratings, we use the root mean squared error
(RMSE) to compare prediction quality.
For PACO, we average predictions over many samples from the posterior, as is
customary in evaluating sampling-based algorithms.

{\bf Perplexity:}
When evaluating the ability to predict held-out reviews, we compare perplexity of review text with other models. 
In order to have a comparable definition of perplexity, following the usual
intuition as the average number of bits necessary to encode a particular word
of review, we transform our model as follows.  We use $\lambda_{u,m}$ as
defined in \eqref{eq:lambda_review} as the vector of expected counts for
each word to be in a review from user $u$ about item $m$. 
We transform this vector to a multinomial with probabilities $\theta_{u,m}$ where
\begin{align}
	\theta_{u,m,x} = \frac{\lambda_{u,m,x}}{\sum_x \lambda_{u,m,x}} \label{eq:poisson_multinomial}
\end{align}
Note that when we average over multiple samples from the posterior, we average
$\lambda_{u,m}$ over those samples and use the averaged rate of the Poisson in
\eqref{eq:poisson_multinomial}.
With this multinomial, we calculate perplexity of testing set $D_{test}$ as 
\begin{align}
	-\log {\rm PPX}(D_{test}) = \frac{1}{N_w} \sum_{(u,m) \in D_{test}} \sum_{x} n_{u,m,x}\log \theta_{u,m,x}
\end{align}
where $N_w$ is the total number of words in held-out testing reviews. Note that this now views our Poisson distributions in their expected state, but
makes our quite different model more easily comparable to previous techniques.

{\bf Joint Negative Log-Likelihood:} 
To evaluate the models jointly in terms of both rating and text prediction, 
we compare the joint negative log-likelihood. Per-review joint negative
log-likelihood is defined as
\begin{align}
	-\log({\rm PPX})\! -\!\frac{1}{\abr{D_{test}}} \!\sum_{(u,m) \in D_{test}}\! \log\left(\Ncal\rbr{\data_{u,m}|\hat{\data}_{u,m}, \sigma^2}\right) \label{eq:joint_likelihood}
\end{align}
The text likelihood is normalized by number of words in review to
equally weight the importance of predicting ratings and text. 
$\sigma$ is taken from the training of each model.

\paragraph{Baseline methods}
We compare PACO with the following models:
\begin{description*}
\item[PMF] Probabilistic matrix factorization (PMF)
	\cite{mnih2007probabilistic} factorizes ratings into latent factors. It is
	simple in structure but usually effective. A number of latent
        factors with Rank $\in
	\cbr{10, 20}$ were tested.\footnote{\label{bpmf_source}Implementation at \url{http://www.cs.toronto.edu/~rsalakhu/BPMF.html} is used in experiments.}
\item[BPMF] Bayesian probabilistic matrix factorization (BPMF)
	\cite{salakhutdinov2008Bayesian} takes a more directly Bayesian approach to
	matrix factorization.
	Number of latent factors Rank $\in \cbr{10, 20}$ were tested.$^{\ref{bpmf_source}}$
\item[HFT] Hidden factors with topics (HFT) \cite{mcauleyhidden} is one of the
	state-of-the-art models that jointly model ratings and reviews. It builds
	connection between topic distribution of reviews and latent factors. It
	shows significant improvement on rating prediction over traditional latent
	factor models on a variety of datasets. We use the implementation from the
	authors' website and run it with parameters recommended in the original paper. 
\item[JMARS] \cite{diao2014jointly} is another state-of-the-art model in
	joint-review-rating modeling.  It explicitly models aspects, sentiments,
	ratings and reviews and provides interpretable and accurate recommendation.
	Similarly, parameters recommended in the original paper is used.\footnote{The
	implementation is in Java.}
\end{description*}
Note, PMF and BPMF only optimize for prediction accuracy on ratings, thus only
focusing on half of the problem we are attacking.  However, we include them for
completeness. 

We test PACO with
different priors and combinations of language models, where we include
per-block language models, per-user cluster language models, or per-item
cluster language models, or some combination thereof.  For all datasets we use
these joint text and rating stencils for the first $S_0$ stencils and then a
series of rating-only stencils.
All results from PACO are reported based for the best RMSE.

\subsection{Quantitative evaluation}

\subsubsection{Joint Predictive Ability}
We compare the ability of \ourmodel to predict jointly reviews and ratings
against that of HFT and JMARS.  In particular, we track the joint negative
log-likelihood by runtime in Figure \ref{fig:joint} and give the detailed
numbers for best results in Table \ref{tab:joint}.
We observe that \ourmodel converges rapidly and has superior
performance to both competitors on all four datasets.
We note that while HFT very quickly reaches reasonable accuracy on both text
and ratings, it very quickly overfits its model of ratings, causing the
surprising curve in Figure \ref{fig:joint}.  
However, results reported in Table \ref{tab:joint} for all models are based on
the best RMSE so as to prevent skew from overfitting.
While we clearly offer high quality joint performance, we now look more closely at
our prediction accuracy for ratings and for text separately.

\begin{table}[t]
\small
\centering
 \begin{tabular}{lllllll}
 \toprule
   &HFT & JMARS & PACO \\
   \midrule
 RateBeer   &52.3546 & 30.2174 & 8.5994 \\
 Yelp  & 20.1377 &    13.3171 & 8.9300 \\
 Amazon fine foods & 14.5827 & 10.1129 & 10.0904\\
 IMDb & 57.4515 & 33.7715 &  31.2567\\
   \bottomrule
 \end{tabular}
 \caption{Joint prediction accuracy for text reviews and ratings, as given by
 joint negative log-likelihood \eqref{eq:joint_likelihood}, for all datasets.  Lower is better.}
 \label{tab:joint}
 \end{table}

\begin{figure}[tbh!]
\centering
\subfigure[Amazon fine foods]{\label{fig:rmse_a}\includegraphics[width=0.32\textwidth]{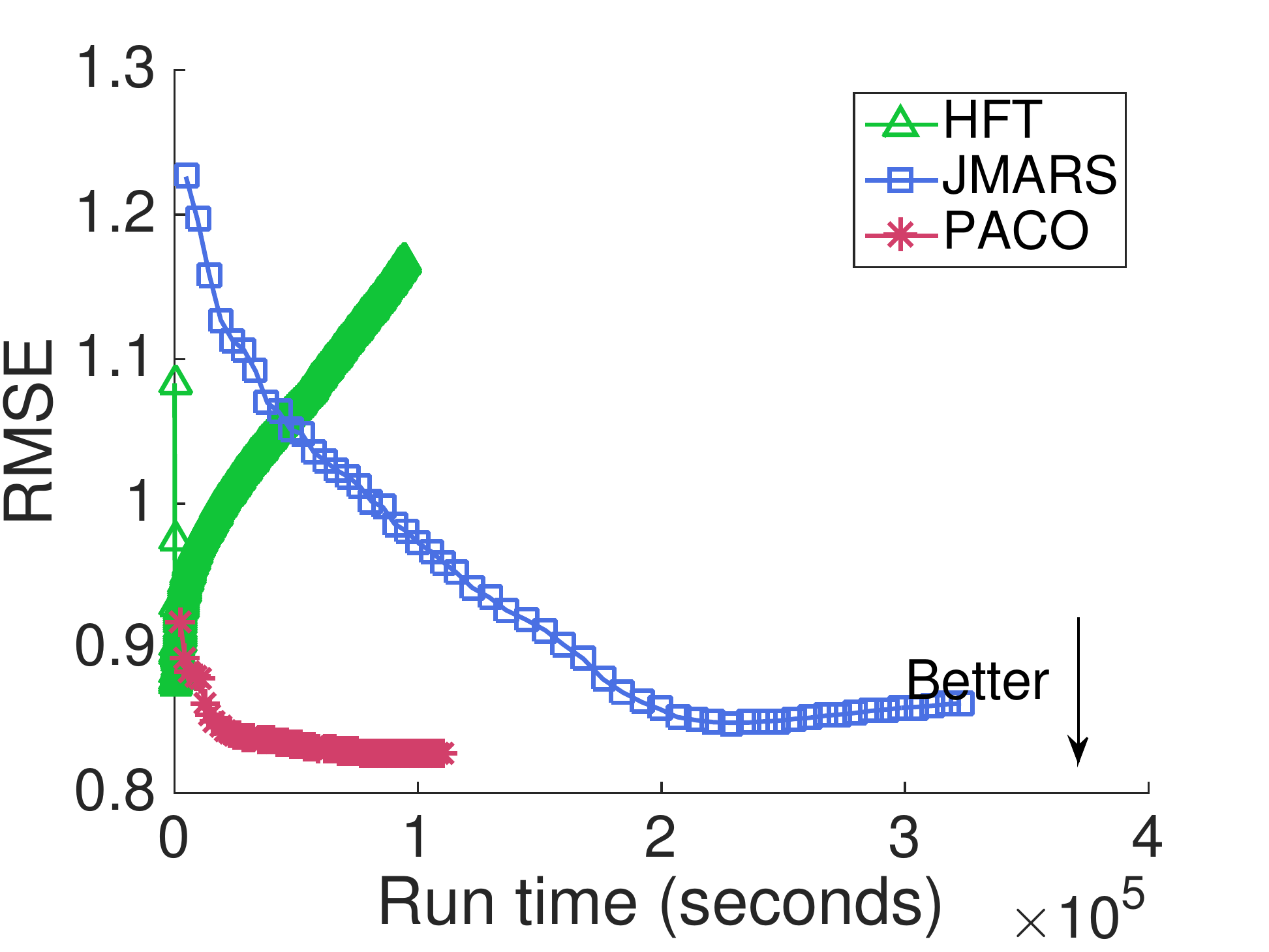}}
\subfigure[RateBeer]{\label{fig:rmse_b}\includegraphics[width=0.32\textwidth]{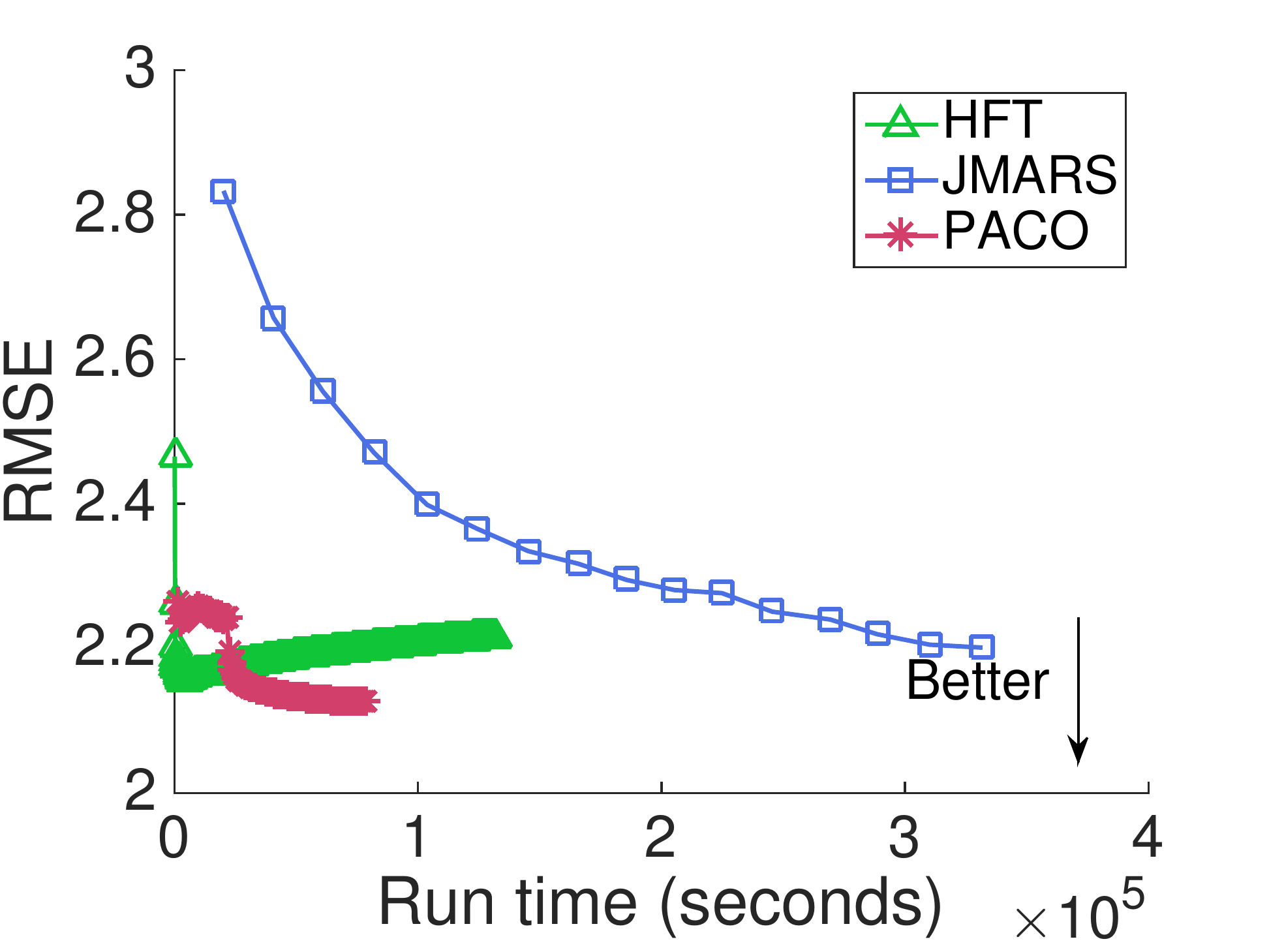}}
\subfigure[Yelp]{\label{fig:rmse_c}\includegraphics[width=0.32\textwidth]{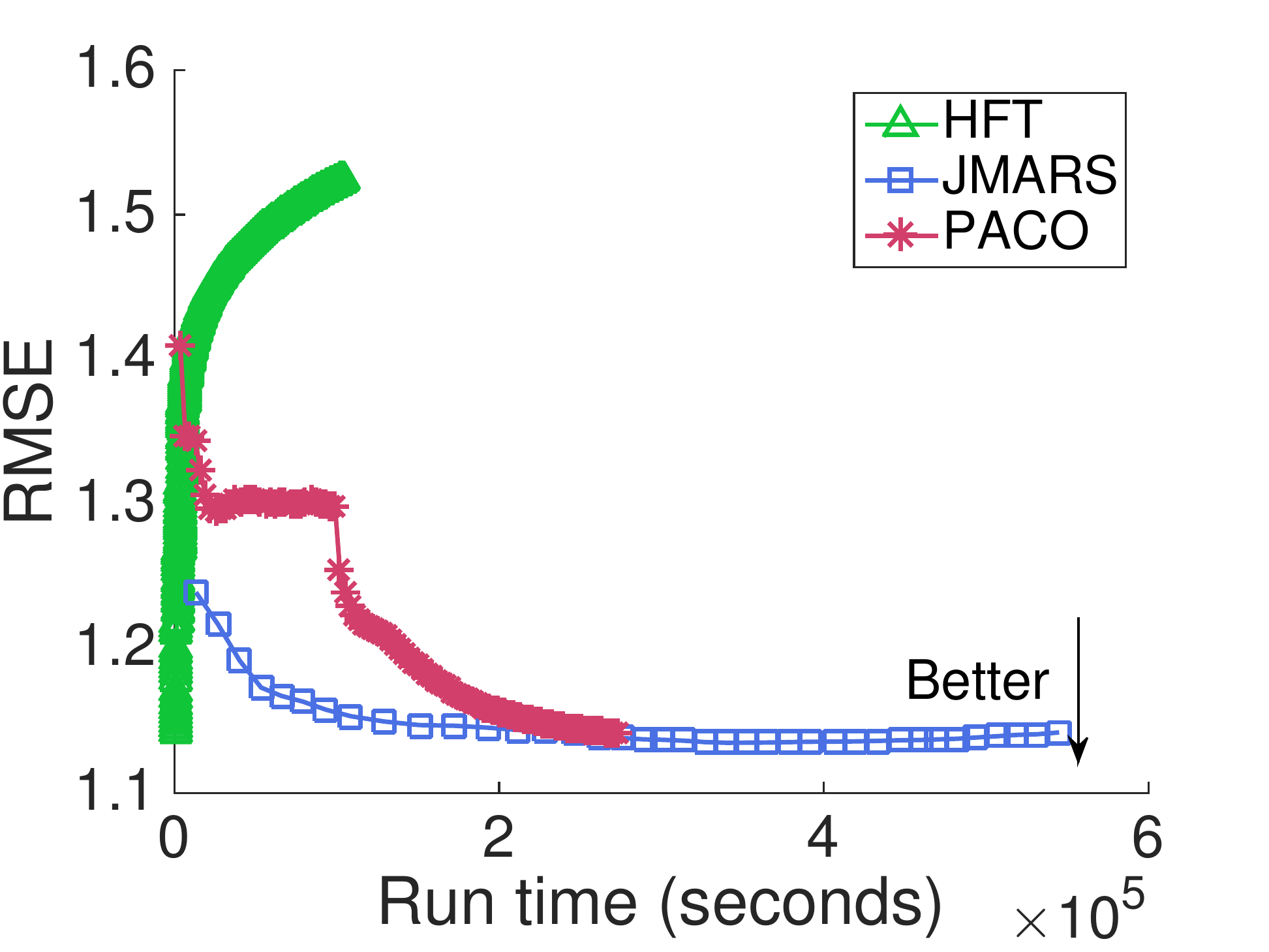}}
\caption{Rating prediction accuracy (RMSE) compared by runtime to other joint
modeling systems$.^4$} 
\label{fig:rmse}
\centering
\subfigure[Amazon fine foods]{\label{fig:ppx_a}\includegraphics[width=0.32\textwidth]{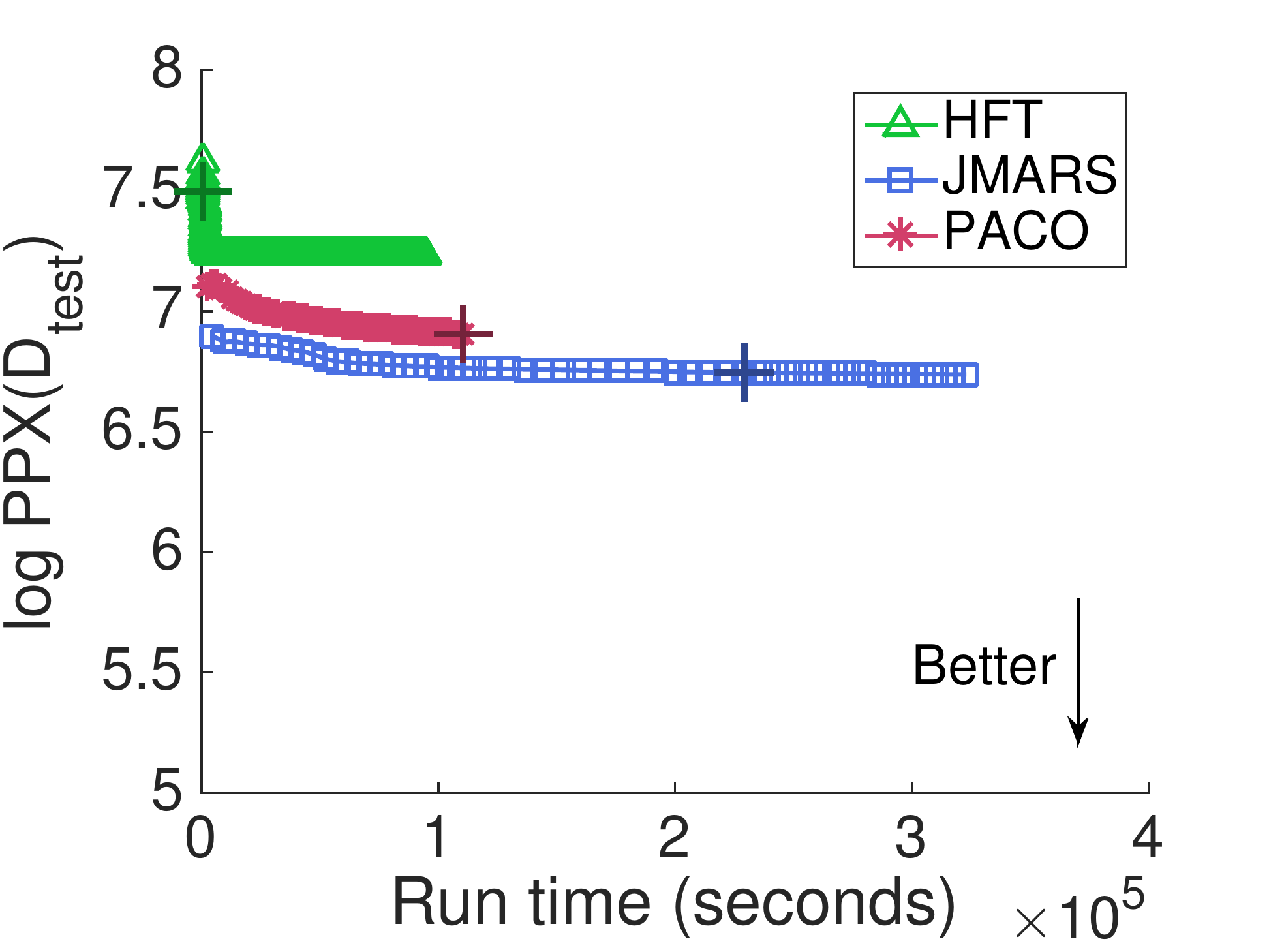}}
\subfigure[RateBeer]{\label{fig:ppx_b}\includegraphics[width=0.32\textwidth]{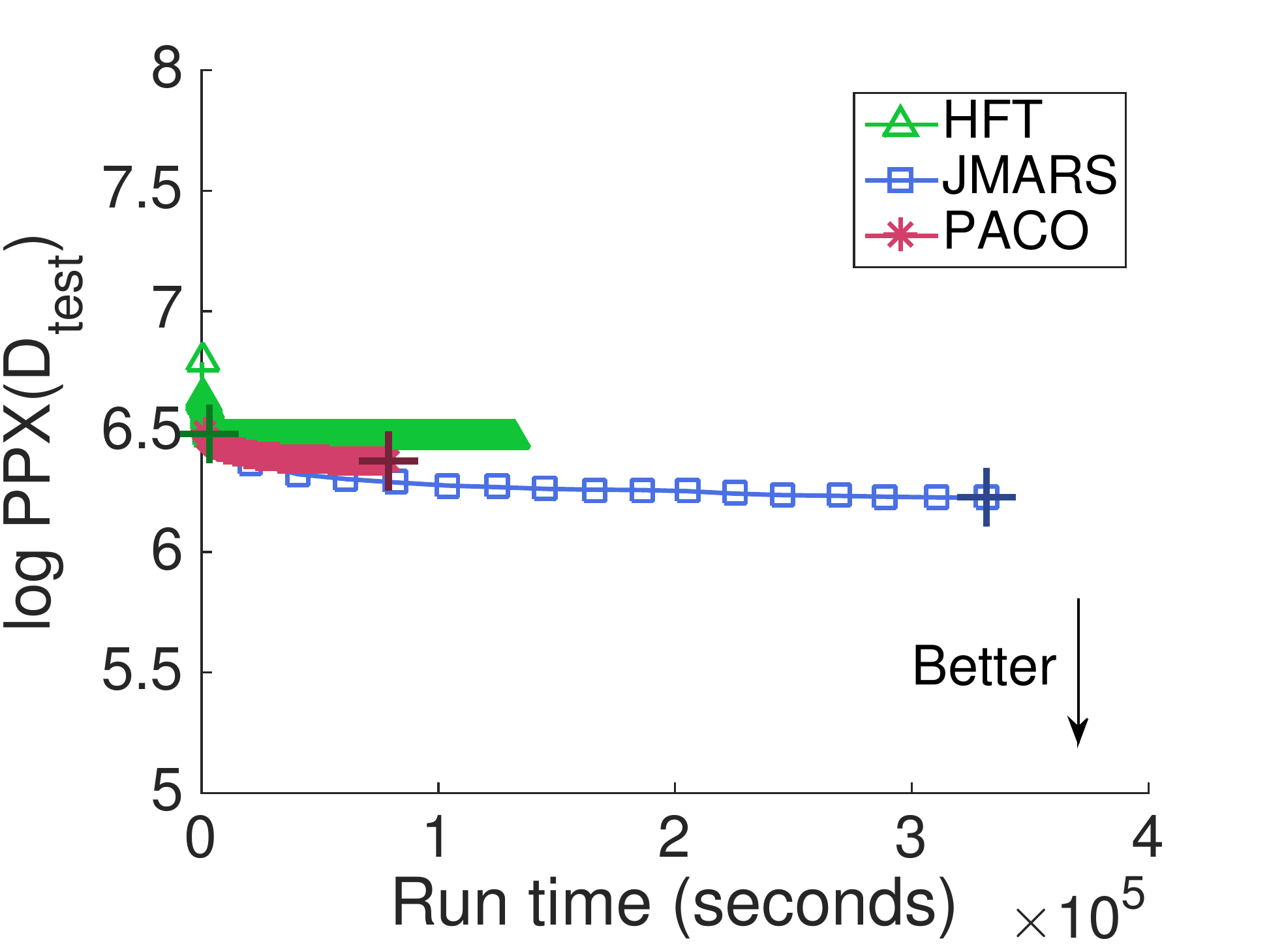}}
\subfigure[Yelp]{\label{fig:ppx_c}\includegraphics[width=0.32\textwidth]{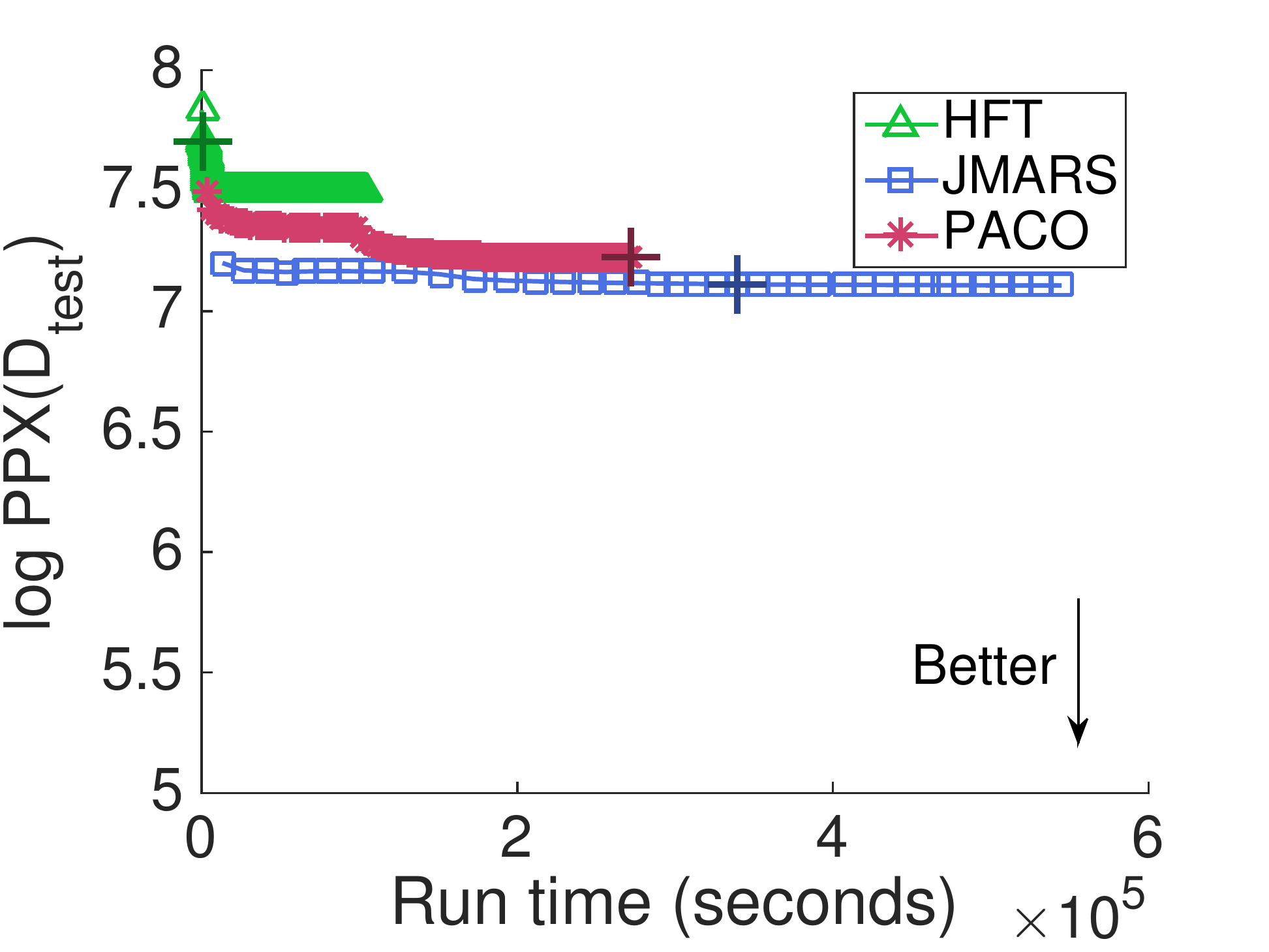}}
\caption{Log perplexity of review text predictions on three datasets. Plus sign markers indicate the values corresponding to the best RMSE. (Lower is better.)} \label{fig:ppx}
\end{figure}

\subsubsection{Rating prediction}
We evaluate performance of rating prediction based on RMSE. Figure
\ref{fig:rmse} shows RMSE over runtime, and Table \ref{tab:rmse} presents
detailed results.
A number of interesting patterns emerge in these results.  
In Figure \ref{fig:rmse}, we observe more clearly that HFT converges extremely
quickly before overfitting, but again results reported in Table \ref{tab:rmse}
are from best RMSE before overfitting.  For PACO we observe reasonable RMSE
before burn-in and then quickly improved RMSE after burn-in.  Finally we
observe that JMARS is generally slower to converge.

In Table \ref{tab:rmse} we observe that \ourmodel has superior performance to
both HFT and JMARS on RateBeer, Amazon fine foods, and IMDb; JMARS performs
slightly better on Yelp.
While \ourmodel generally outperforms
the competing joint models, in these experiments it achieves slightly worse
accuracy than BPMF. 
Here we observe a general trade-off between size and performance. 
As discussed in \cite{accams}, using a sum of co-clusterings is far more succinct than
bilinear models like BPMF.  As a result, the ratings model in \ourmodel is far
smaller than that of BPMF, while we consistently achieve nearly as high of an
accuracy.

\begin{table}[t!]
\small
\centering
 \begin{tabular}{lllllll}
 \toprule
   & PMF & BPMF &HFT & JMARS & \ourmodel \\
   \midrule
 RateBeer   &2.1944& 2.1164 & 2.1552 & 2.1675 & 2.1273\\
 Yelp  & 1.2649 &1.1346  & 1.1408 & 1.1347  & 1.1407 \\
 Amazon fine foods &0.8752 & 0.8193  &0.8809 & 0.8486  & 0.8292\\
 IMDb & 2.5274& 2.1622 & 2.2328& 2.2947 & 2.1877\\
   \bottomrule
 \end{tabular}
 \caption{Rating prediction accuracy (RMSE) across all four datasets.  We
 observe that \ourmodel generally outperforms the other joint learning models (HFT and
 JMARS) as well as PMF.}
 \label{tab:rmse}
 \end{table}

\subsubsection{Review Text Prediction}
The second component of the predictive ability of our model is predicting
review text, as measured by perplexity. 
We give the perplexity over runtime in Figure \ref{fig:ppx}.
We observe an apparent trade-off in perplexity for speed, simplicity and accuracy in
rating prediction.  PACO is efficient and outperforms HFT on all datasets.
Note, in \cite{mcauleyhidden}, HFT is described to primarily use text to
improve rating predictions and not for predicting review.
JMARS gives slightly better perplexity at the cost of significantly more complex model. 
Precise perplexity of HFT, JMARS, and \ourmodel are given in Table \ref{tab:ppx}. 
Since our primary goal is recommendation, the perplexity reported correspond to
the points that obtain the best RMSE. 

\begin{table}[tbh]
\small
\centering
 \begin{tabular}{llllll}
 \toprule
 & HFT & JMARS & PACO \\
   \midrule
 RateBeer  & 6.4891 & 6.2098  & 6.3779\\
 Yelp   &7.7031 & 7.1112  & 7.2223 \\ 
 Amazon fine foods   &7.5015 & 6.7450 & 6.8759 \\
 IMDb   &8.1747 & 7.4610 & 7.5540  \\ 
   \bottomrule
 \end{tabular}
 \caption{Perplexity on all four datasets. The perplexity values reported here correspond to the points that obtain the best RMSE. }\label{tab:ppx}
 \end{table}
  
\subsubsection{Cold-start}
\begin{figure}[tbh]
\centering
\subfigure[Amazon fine foods]{\label{fig:joint_a}\includegraphics[width=0.32\textwidth]{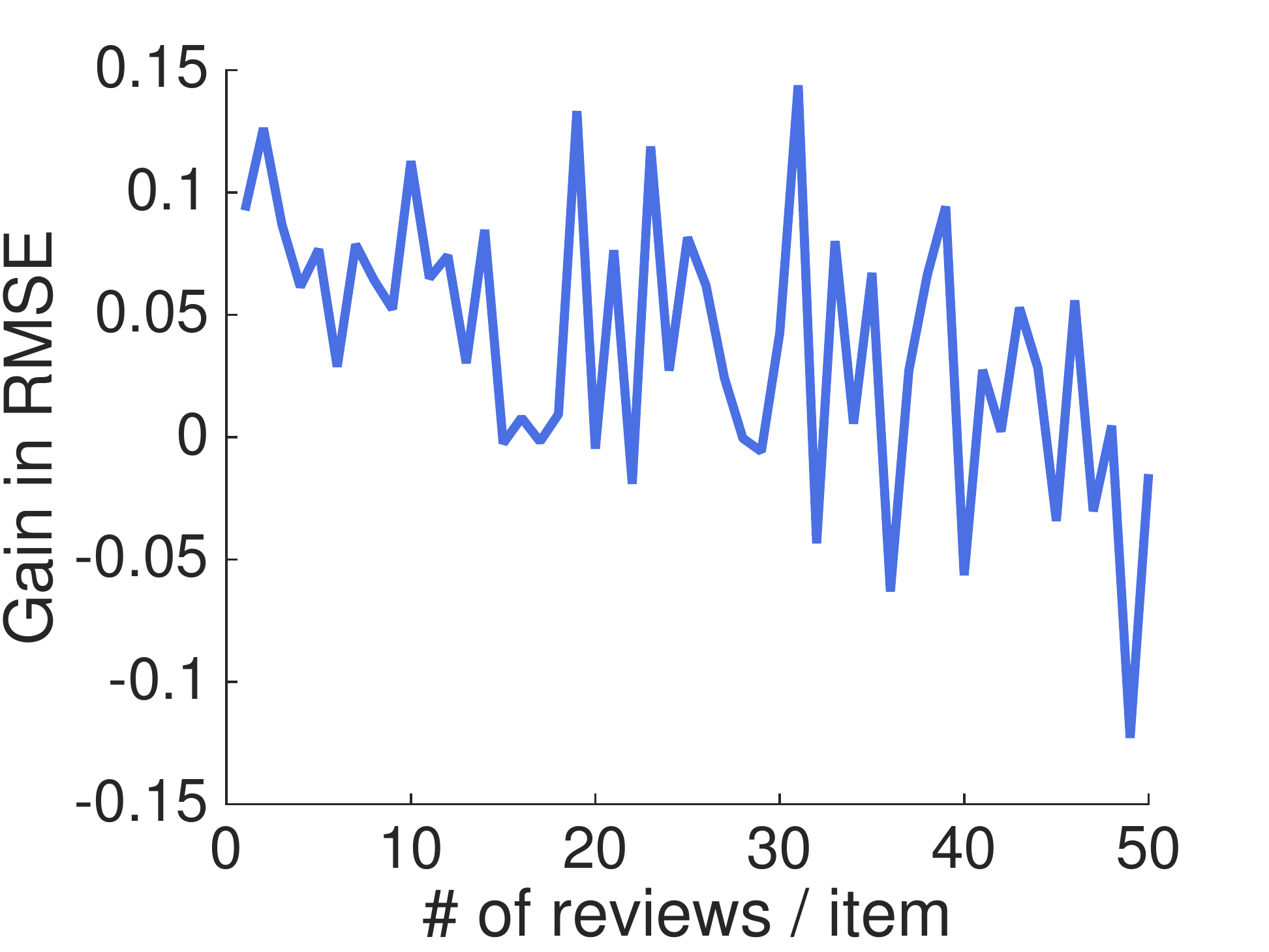}}
\subfigure[RateBeer]{\label{fig:joint_b}\includegraphics[width=0.32\textwidth]{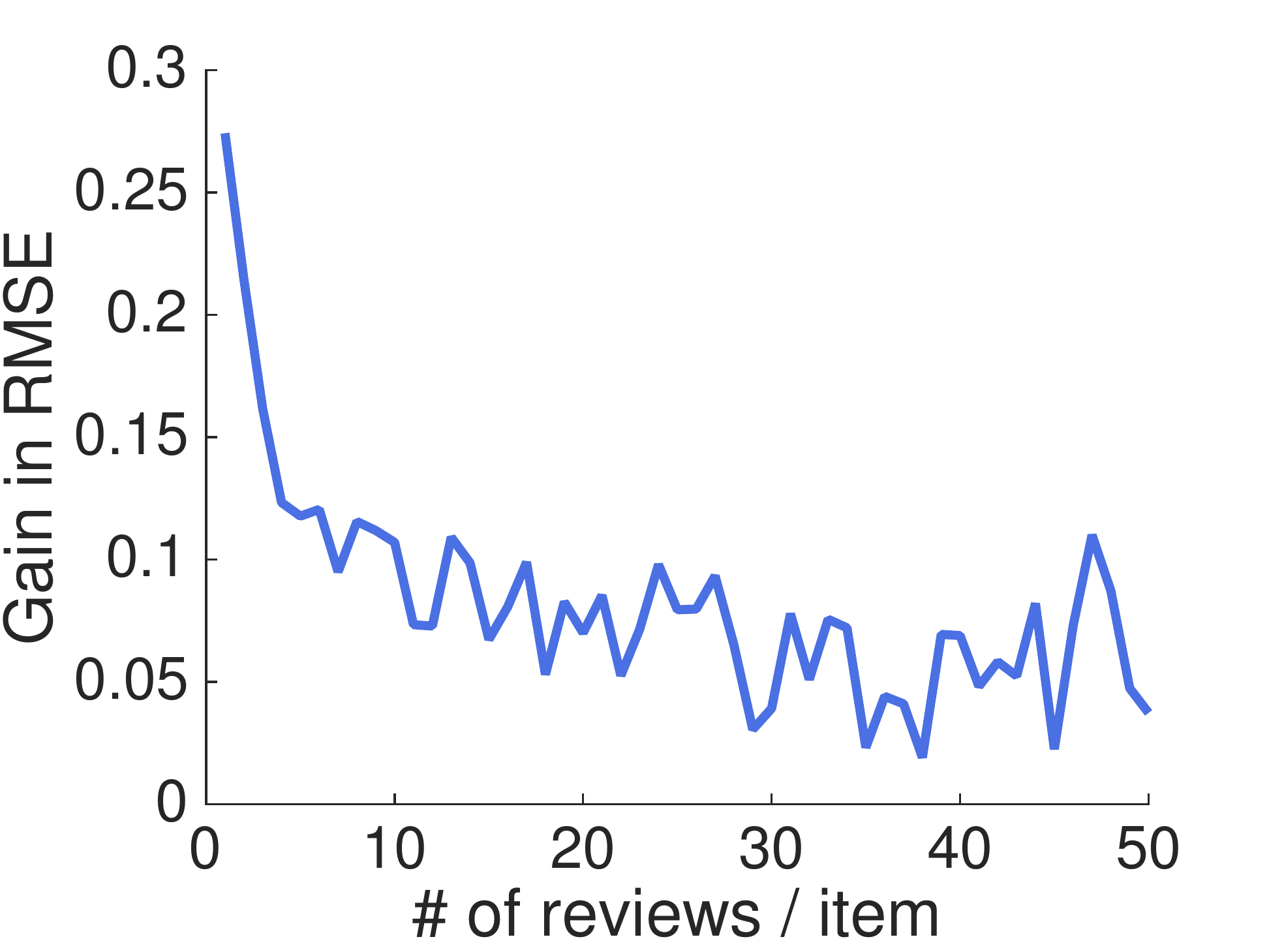}}
\subfigure[Yelp]{\label{fig:joint_c}\includegraphics[width=0.32\textwidth]{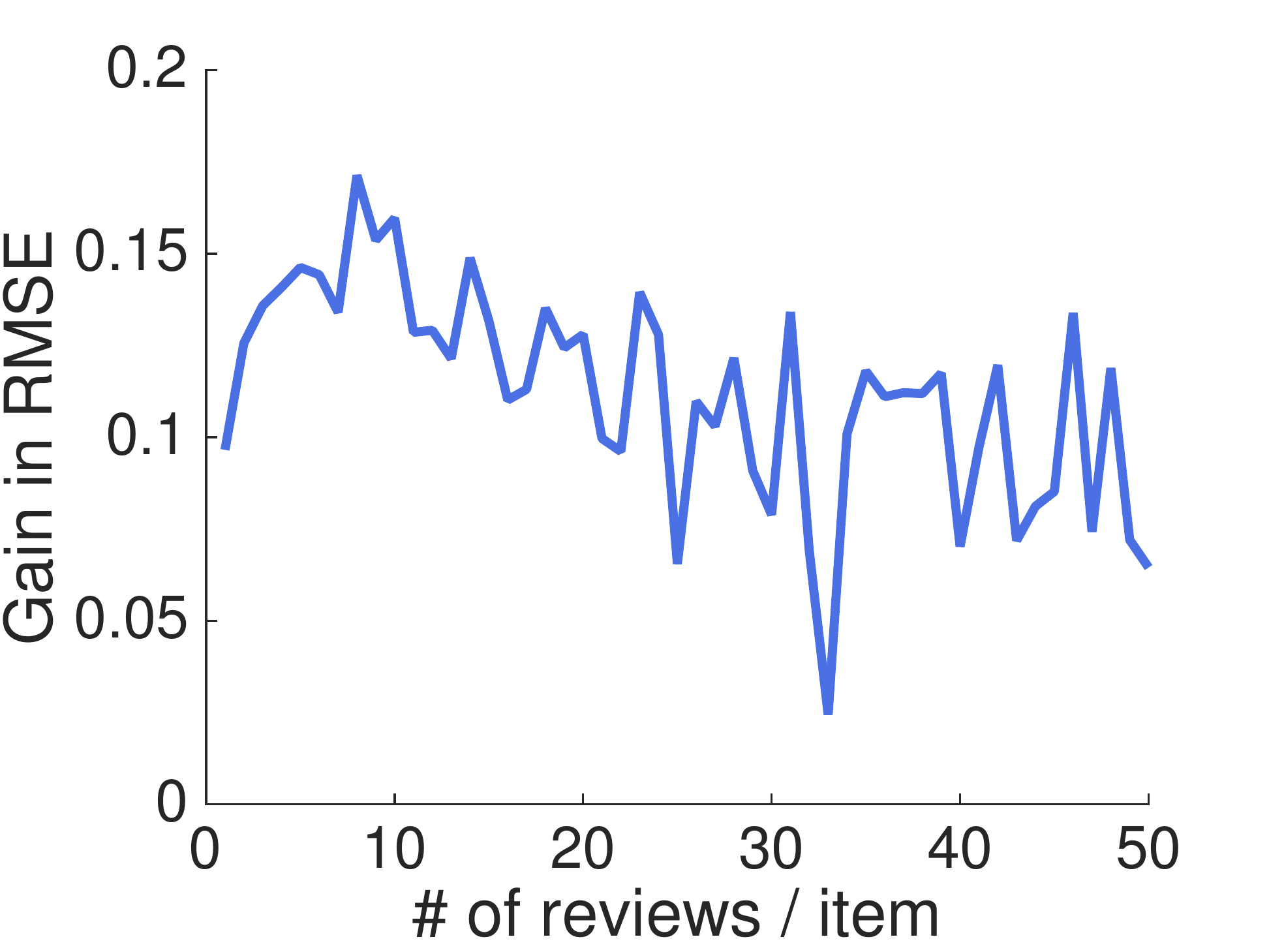}}
\caption{Gain of \ourmodel over PMF in RMSE demonstrates the benefits for items with few observed ratings. } \label{fig:cold_item}
\centering
\subfigure[Amazon fine foods]{\label{fig:joint_a}\includegraphics[width=0.32\textwidth]{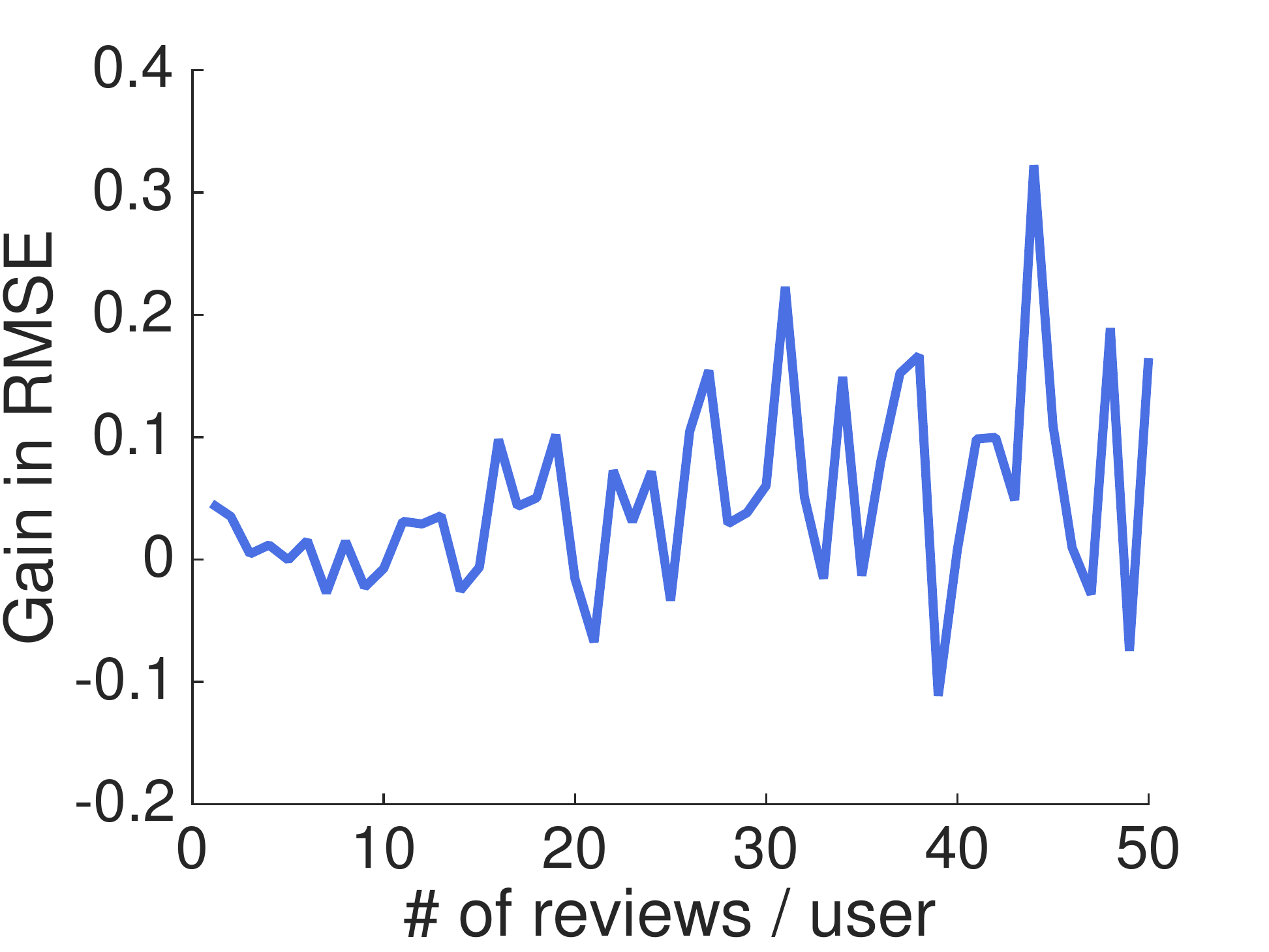}}
\subfigure[RateBeer]{\label{fig:joint_b}\includegraphics[width=0.32\textwidth]{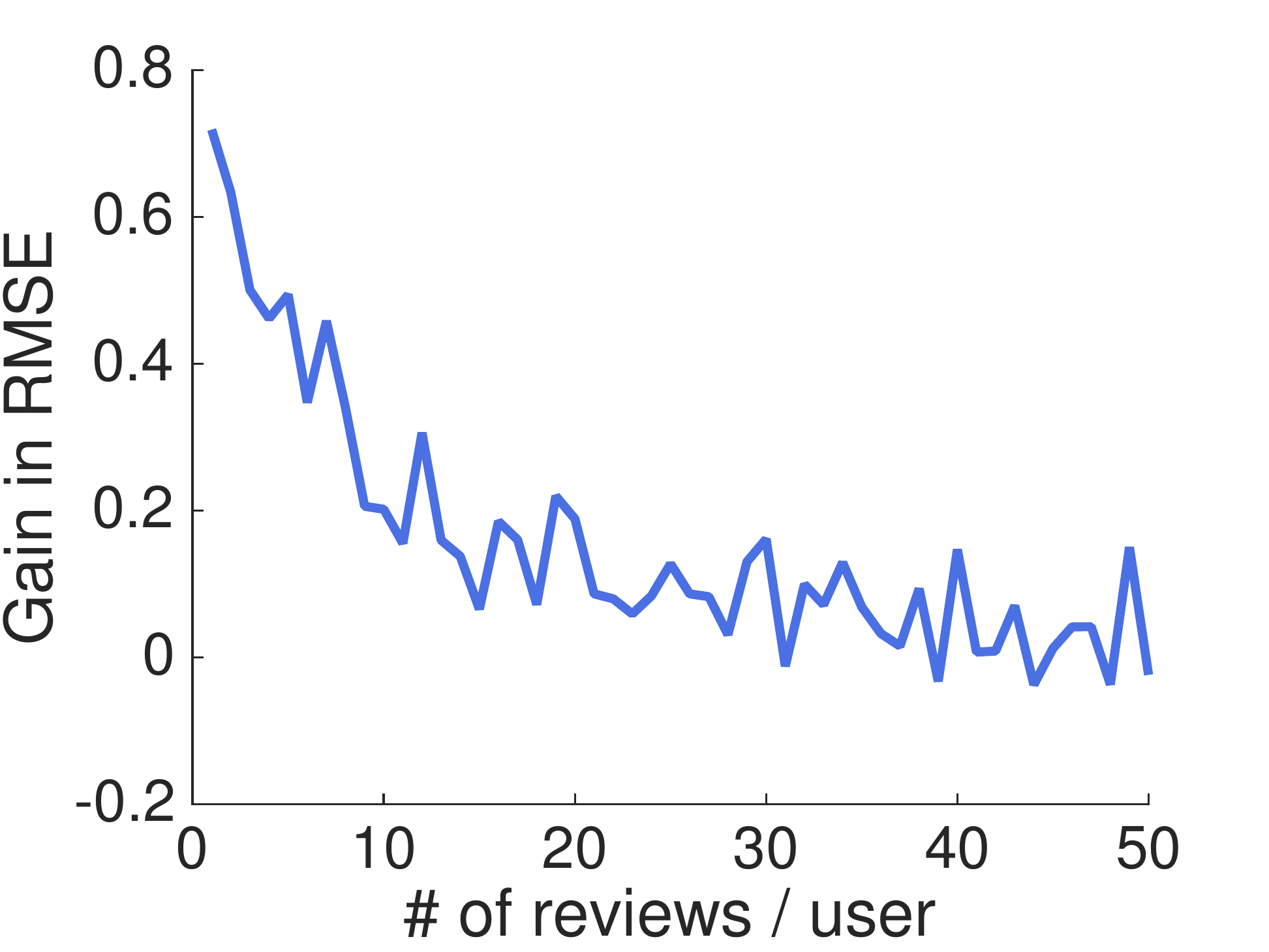}}
\subfigure[Yelp]{\label{fig:joint_c}\includegraphics[width=0.32\textwidth]{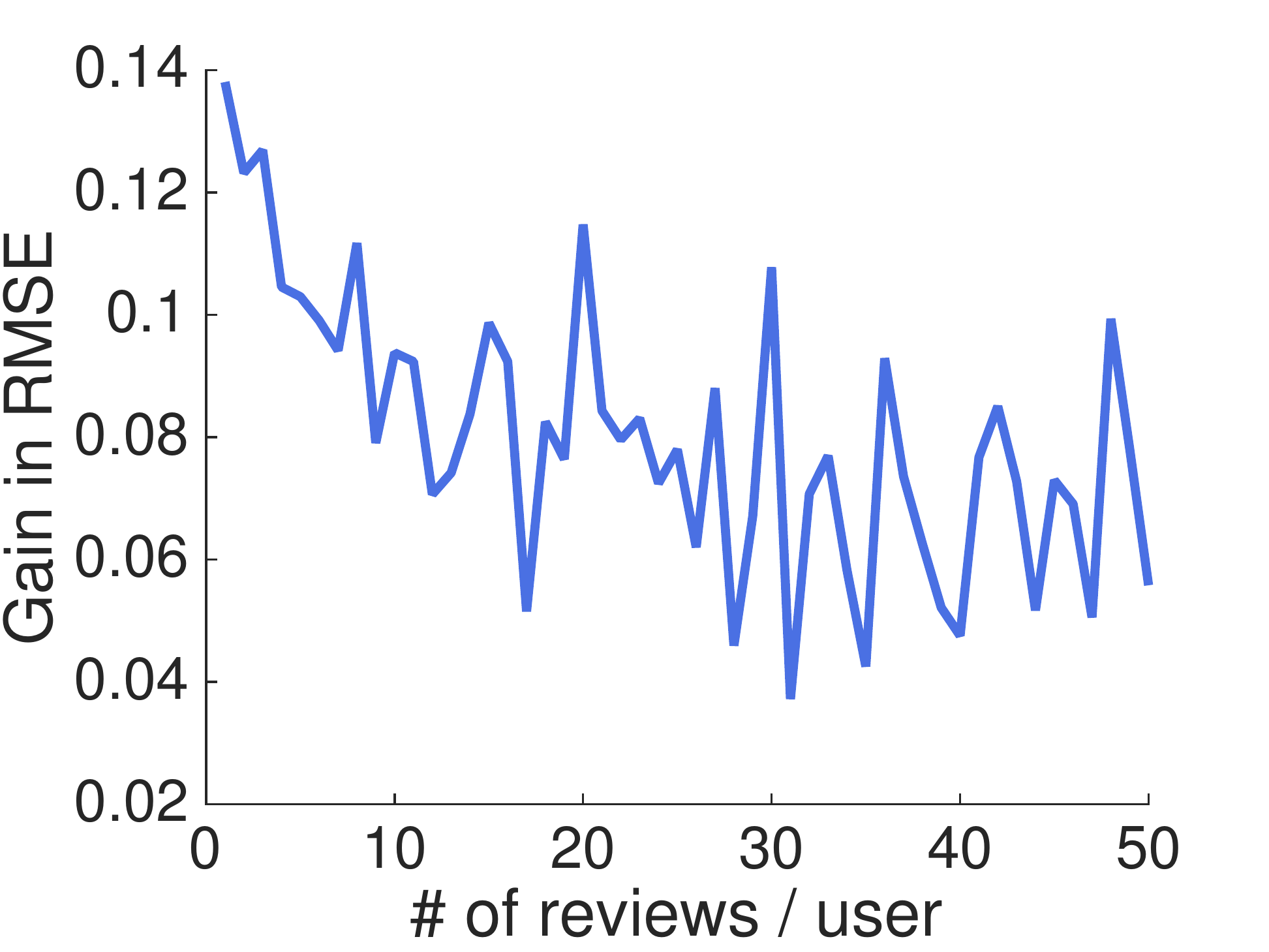}}
\caption{Gain of \ourmodel over PMF in RMSE demonstrates the benefits for users with few observed ratings. } \label{fig:cold_user}
\end{figure}

For the sake of thoroughness, we looked at where does our joint model excel
at rating prediction over the rating-only PMF model.
We generally found PACO to help alleviate cold-start challenges.
We show improvement in RMSE over PMF for
items and users with different number of training examples in Figure
\ref{fig:cold_item} and \ref{fig:cold_user}. 
We observe that for items with fewer observed ratings we achieve a
greater improvement in rating prediction accuracy.
This suggests that PACO is able
to extract rich information from reviews and provide benefits especially
when items have scarce signals. 
For users with few ratings, we can see similar trends except for Amazon fine foods.
One hypothesis is that the quality of food is 
less subjective than movies or beers. It is thus harder to learn user
preference from the review in this case. 

\subsection{Interpretability}
In addition to quantitatively evaluating our method, we also want to
empirically demonstrate that the patterns surfaced and review predictions would
be useful to the human eye.  
However, because reviews from each dataset follow very different patterns, we
expect \ourmodel to model them differently.
In this effort, we analyzed our learned models to understand which parts
of the models were effective in understanding each of these datasets.

\noindent \textbf{Sentiment word extraction:}
We first checked to make sure that blocks in our co-clustering that had high
rating predictions found appropriately positive or negative words.
In Table \ref{tab:sentiment} we present the top words in the highest-rating
block and lowest-rating block in the first stencil for each dataset. 
We can see in general, in that high-rating blocks, there are more positive sentiment
words, and in lower-rating blocks, more negative words. 
This is less clear in the RateBeer and Amazon fine foods datasets where reviews more strongly focus on
describing the item. However, we still note the strong association with
``good'' for a positive rating and in some cases ``bad'' for a negative rating.
\begin{table*}[th]
\scriptsize
\centering
 \begin{tabular}{lll}
 \toprule
 Dataset& Rating& Words\\
 \midrule
 \multirow{ 2}{*}{IMDb} &0.022&great, love, movies, story, life, watch, time, people, character, characters, best, films, scene, watching\\
 &-0.018&bad, good, plot, like, worst, money, waste, acting, script, movies, minutes, horrible, boring, thought\\
 \midrule
 \multirow{ 2}{*}{Fine foods} &0.024&good, love, great, like, product, amazon, store, time, find, eat, price, years, buy, coffee, taste, stores\\
&-0.048 & like, buy, taste, time, product, bought, purchased, good, pretty, thought, reviews, smell, purchase\\
 \midrule
 \multirow{ 2}{*}{Yelp} &0.51&highly, recommend, professional, amazing, job, customer, work, best, appointment, staff, great, needed\\
&-0.76&told, manager, customer, called, call, rude, asked, horrible, worst, phone, minutes, order, money, hotel\\
 \midrule
 \multirow{ 2}{*}{RateBeer} &0.15&nice, pours, hops, flavor, hop, citrus, color, taste, finish, tap, good, sweet, bitterness, white, malt, light\\
&-0.22&taste, bad, beer, like, color, good, decent, weak, drink, pours, beers, boring, special, watery, dont\\
 \bottomrule
 \end{tabular}
 \caption{Blocks predict words matching the sentiment of the predicted rating.}
 \label{tab:sentiment}
 \end{table*}
 
\noindent \textbf{Item-specific words:}
In Table \ref{tab:specific}, top item-specific words associated with some
popular items are presented. In general these words provide basic descriptions unique 
to that particular item. 
We do observe some overfitting in cases where there are fewer reviews, but
generally the item-specific language model improves predictive performance. 
\begin{table*}[!tbh]
\scriptsize
\centering
 \begin{tabular}{ll}
 \toprule
 Item & Item-specific words\\
 \midrule
The Dark Knight & batman, joker, dark, ledger, knight, heath, nolan, best, performance, bale\\
Silent Hill & game, silent, hill, games, horror, video, rose, town, like, played, plot, scary\\
 \midrule
 Canned Kitten Chicken Liver Food & food, time, kittens, kitty, liver, cats, feral, white, plan, female, guys, kitten\\
 Low Fat Clusters and Flakes Cereal Raisin Bran & cereal, sugar, fat, wheat, calories, color, barley, raisins, ingredients, listed\\
 \midrule
 Luxury Nail \& Spa & hair, appointment, stylist, cut, salon, desk, paying, stylists, bid, grille, color\\
 Enrico's Tazza D'oro Cafe \& Espresso Bar&coffee, espresso, pittsburgh, pastries, park, baristas, cappuccino, atmosphere\\
 \midrule
 Barley Island Sheet Metal Blonde & wheat, orange, wit, coriander, clove, bubblegum, slice, chamomile, witbier\\
Rock Bottom Braintree Boston Fog Lager&peaches, dishwater, lager, perfumey, dissipating, fog, outspoken, component\\
  \bottomrule
 \end{tabular}
 \caption{Item-specific words capture concepts highly specific to the individual item.} 
 \label{tab:specific}
 \end{table*}

\noindent\textbf{Item-clusters and cluster-specific words:}
In Table \ref{tab:cluster} we present 4 clusters of items and the words associated with
them from Amazon fine foods and IMDb. 
For Amazon fine foods, we see items with similar categories are clustered. 
For example, the coffee cluster and the snack cluster are learned effectively, as presented in the first and second row of the table.
On the other hand, PACO learns relatively general clusters for IMDb. For example, while one
presented cluster captures exciting or action movies, the other one groups
generally lower-rated movies, and the associated words are general
negative sentiment words that would be in a movie review. 

\begin{table*}[!tbh]
\scriptsize
\centering
 \begin{tabular}{p{8cm}|p{8cm}}
 \toprule
 Subset of items in cluster & Cluster words\\
\midrule
 Melitta Cafe de Europa Gourmet Coffee, Flavored, Coffee People Black Tiger, Dark Roast, K-Cup for Keurig Brewers, K-Cup Portion Pack for Keurig K-Cup Brewers, Lavazza Super Crema Espresso - Whole Bean Coffee, Starbucks Sumatra Dark, K-Cup Portion Pack for Keurig K-Cup Brewers (Amazon fine foods) & like, coffee, good, taste, flavor, cup, drink, nice, product, thought, great, tastes, tasting, drinking, best, full, time, buy, recommend, enjoy, brand, love, strong, blend, black, regular, bit, bad, recommended, size \\
 \midrule
 Ice Breakers Ice Cubes Sugar Free Gum, Kiwi Watermelon, Bell Plantation PB2 Powdered Peanut Butter, PB2 Powdered Peanut Butter, Ella's Kitchen Organic Smoothie Fruits, The Red One, Blue Diamond Almonds Bold Lime n Chili, Blue Diamond Almonds Wasabi \& Soy Sauce, Value Pack (Amazon fine foods)& taste, snack, like, good, eat, eating, buy, bag, love, diet, healthy, fat, great, flavor, store, sweet, salty, healthier, price, amount, bags, find, case, crunchy, size, tasty, ate, packs, texture, yummy \\
\midrule
Entrapment, Mission: Impossible III, Zombie, Snake Eyes, Starsky \& Hutch, New England Patriots vs. Minnesota Vikings, I Am Legend, Chaos (IMDb)& action, good, character, thought, story, plot, scene, expected, average, movies, game, scenes, lack, massive, destruction, entertained, suspenseful, audience, seats, batman, pulls, mistakes, steel, effect, shopping, richardson, atmosphere, ford, genetic, horrific \\
\midrule 
Gargantua, Random Hearts, Chocolate: Deep Dark Secrets, Blackout, The Ventures of Marguerite, Irresistible, Ghosts of Girlfriends Past, Youth Without Youth  (IMDb)& like, good, bad, time, movies, people, acting, plot, watch, horror, watching, worst, scenes, pretty, awful, effects, scene, characters, thought, story, actors, worse, films, terrible, special, lot, fun, give, stupid, guy\\

  \bottomrule
 \end{tabular}
 \caption{Discovered clusters of items and associated topics for Amazon fine foods and IMDb.}
 \label{tab:cluster}
 \end{table*}

\begin{table*}[!tbh]
\scriptsize
\centering
 \begin{tabular}{p{8cm}|p{8cm}}
 \toprule
 {Real review} & {Predicted words (ordered by likelihood)} \\ 
 \midrule
poured from the bottle pitch black with a caramel head smells like a great espresso with a little bit of oatmeal in there great creamy mouthfeel tastes is strong of very bitter coffee and oatmeal the booze is pretty well hidden this is one tasty stout
 & 
coffee, head, aroma, beer, roasted, sweet, light, malt, bitter, bottle, taste, stout, flavor, dark, like, finish, thick, white, brown, nice, creamy, good, tan, medium, pours, smooth, chocolate, body, caramel, great 
\\ 
\midrule
tap at pour is hazy orange gold with a white head aroma shows notes of wheat tangerine orange yeast and coriander flavor shows the same with light vanilla
& 
orange, white, head, citrus, aroma, light, wheat, sweet, hazy, flavor, malt, yeast, finish, beer, coriander, spice, medium, bottle, body, nice, taste, hops, good, lemon, pours, cloudy, bitter, notes, color, caramel
\\ 
\midrule
this is a pale ale it is a pale orangish color and it is an ale hops dominate the nose but there is a more than ample malt backbone good medium mouthfeel clean hoppy follow through beer as it ought to be
&
head, hops, aroma, ipa, nice, good, flavor, beer, citrus, hop, hoppy, taste, sweet, finish, bottle, malt, white, pours, light, color, medium, pine, golden, bitter, like, grapefruit, body, amber, floral, caramel
\\ 
\bottomrule
 \end{tabular}
 \caption{Examples of predicted words for held-out reviews on RateBeer.}
 \end{table*}

\noindent\textbf{Review prediction:}
Finally, for RateBeer we give examples of generally well-predicted held-out
reviews and the top words the entire \ourmodel model predicts for them.
Because RateBeer reviews are largely descriptive of the item, we find that
\ourmodel is effective in predicting the properties of the items, particularly
focused on the type of beer.

\section{Conclusion}
\label{sec:Conclusion}

We presented \ourmodel, an additive co-clustering algorithm for
explainable recommendations. The key goal of this work was to
demonstrate that the additive co-clustering approach proposed in
ACCAMS is highly versatile and can be extended to texts. As a useful
side-effect we obtained an \emph{additive} formulation for language
models, complete with an efficient sampler. This technique may be
useful in its own right. 

Empirical evidence demonstrates that \ourmodel models both ratings and
text well on a variety of datasets. In particular, it is able to
extract attributes of items, users, specific clusters, or attached to
sentiments. This versatility allows one to go beyond gazetteered
collections of words and use highly flexible and autonomous models
that can be applied with very little knowledge of the specific
language, that is, they are very useful for the purpose of
internationalization of recommender systems that are able to
understand a user's opinions.



\renewcommand{\refname}{\normalfont\selectfont\bfseries{\Large References}}
\bibliographystyle{plain}

\end{document}